\documentclass[twocolumn]{article}
\usepackage[margin=0.5in]{geometry}

\usepackage{framed,multirow}

\usepackage{amssymb}
\usepackage{latexsym}

\usepackage{url}
\usepackage{xcolor}
\usepackage{hyperref}
\usepackage{caption}
\usepackage{subcaption}
\usepackage{graphicx}
\usepackage{float}
\usepackage{fancyhdr}

\definecolor{newcolor}{rgb}{.8,.349,.1}
\graphicspath{{images/}, {pdf-images/}}

\fancypagestyle{plain}{%
\fancyhead{}
\fancyhead[c]{\textit{Published at CVIU: \url{https://doi.org/10.1016/j.cviu.2024.103934}}}
}
\begin{document}

\title{On the Coherency of Quantitative Evaluation of Visual Explanations}
\author{Benjamin Vandersmissen\thanks{Primary author: \href{mailto:benjamin.vandersmissen@uantwerpen.be}{benjamin.vandersmissen@uantwerpen.be}} \qquad Jos\'e Oramas \\ \\ IDLab, University of Antwerp - imec \\ Prinsstraat 13, 2000 Antwerp, Belgium}
\date{October 2023}

\maketitle

\begin{abstract}
Recent years have shown an increased development of methods for justifying the predictions of neural networks through visual explanations. These explanations usually take the form of heatmaps which assign a saliency (or relevance) value to each pixel of the input image that expresses how relevant the pixel is for the prediction of a label.
Complementing this development, evaluation methods have been proposed to assess the "goodness" of such explanations. 
On the one hand, some of these methods rely on synthetic datasets. However, this introduces the weakness of having limited guarantees regarding their applicability on more realistic settings. 
On the other hand, some methods rely on metrics for objective evaluation. However the level to which some of these evaluation methods perform with respect to each other is uncertain.
Taking this into account, we conduct a comprehensive study on a subset of the ImageNet-1k validation set where we evaluate a number of different commonly-used explanation methods following a set of evaluation methods. We complement our study with sanity checks on the studied evaluation methods as a means to investigate their reliability and the impact of characteristics of the explanations on the evaluation methods.
Results of our study suggest that there is a lack of coherency on the grading provided by some of the considered evaluation methods. Moreover, we have identified some characteristics of the explanations, e.g. sparsity, which can have a significant effect on the performance.
\end{abstract}


\section{Introduction}
Neural Network models have risen from obscurity to the State-of-the-Art in computer vision tasks such as Visual Question Answering (VQA)~\cite{antol2015vqa, wu2017visual}, Image Segmentation~\cite{garcia2017review}, Image Classification~\cite{krizhevsky2012imagenet}, Object Localisation~\cite{tompson2015efficient}, Image Captioning~\cite{you2016image, vinyals2016show} and more. This explosive rise in effectiveness and popularity can be attributed to better, larger datasets such as ImageNet~\cite{deng2009imagenet, russakovsky2015imagenet}, and Pascal-VOC~\cite{everingham2010pascal}, as well as exponentially increasing compute power via stronger, more specialised hardware.

As neural network models are more and more ubiquitous and are applied in critical real-world applications, such as medicine~\cite{hosny2018artificial}, self-driving cars~\cite{jheylenWACVautonomousDrivingCNN,ni2020survey}, military applications~\cite{svenmarck2018possibilities} and more, the need arises for explainability. End-users should no longer blindly accept the predictions of a neural network model, but rather the model should try to justify its predictions towards the user. In doing so, the user can determine via expert knowledge in the domain whether the reasoning of the model is sound and valid.


Common tools used in this context are visual explanation methods, algorithms that assign relevance scores to features in the input image based on how relevant the feature is to the predicted label. Numerous explanation methods have been introduced in the literature, including but not limited to: Grad-CAM~\cite{selvaraju2017grad}, Guided Backpropagation~\cite{springenberg2015striving}, Layerwise Relevance Propagation~\cite{bach2015pixel, montavon2019layer}, LIME \cite{ribeiro2016should}, RISE~\cite{petsiuk2018rise}, Sliding-Window Occlusion~\cite{zeiler2014visualizing}, Excitation Backprop~\cite{zhang2018top} and SHAP-values~\cite{lundberg2017unified}.

Previous work evaluating multiple visual explanation methods can be classified in two categories: large-scale studies such as \cite{tjoa2020quantifying, arras2022clevr} that use computer-generated data to compare the effectiveness of a large number of different methods, and papers that propose a new explanation method and compare their proposed method to a small set of directly related methods. As far as our knowledge, no large scale comparison has been done using standard datasets depicting real-world scenes. 
Taking the above into account, the contributions of this paper are: i) a large-scale comparison and evaluation of multiple commonly used explanation methods in a more realistic scenario, as well as ii) an in-depth study of the most common evaluation methods for visual explanations.

The rest of this paper is structured as follows. \autoref{sec:related} presents an overview of the related work, while \autoref{sec:methodology} will lay down the experimental methodology followed during the study. In \autoref{sec:experiments} we share our experiments and results, and we provide a high level discussion in \autoref{sec:discussion}. Finally, we conclude this paper in \autoref{sec:conclusion}.

\section{Related Work}
\label{sec:related}

\subsection{Explainability vs Interpretability}

In the literature, the terms interpretability and explainability are sometimes used interchangeably. However this is not entirely correct, as these terms are subtly distinct in meaning.
\textit{Explainability} is instance specific justification, justifying \emph{why a specific prediction was made}. It involves generating an explanation for a specific input and  prediction, such that the explanation highlights the features present in the given input that determine the prediction of the given label. In the case of visual explanations, an explanation is given by a saliency map that assigns an importance value to each pixel in the image.
\textit{Interpretability} is model-wide justification, indicating which features the model has learned and how it associates those features with the predictions it is trained to generate. The end goal of Interpretability is aligning neural network learned features with human interpretable concepts, concepts that are easy to understand for humans. This is important to better understand which relationships the neural network model has learned and which internal representations are used.
Here we focus our analysis on explanation methods.

\subsection{Explanation Methods}

The explanation methods introduced in the literature can be roughly grouped in four main categories: perturbation-based methods, gradient-based methods, CAM-based methods and backpropagation-based methods.

\textbf{Perturbation-based methods}
generate visual explanations by repeatedly perturbing the input image and using the prediction score difference between the perturbed image and the original image as a measure of relevance for the perturbed area. This has the side effect that these approaches do not need access to neural network internals and as such can also generate explanations for other types of models. These methods include, but are not limited to, sliding window occlusion~\cite{zeiler2014visualizing} and RISE~\cite{petsiuk2018rise}.

\textbf{Gradient-based methods} use the gradient of the input w.r.t. the prediction score to explain neural network model decisions. Originally pioneered in \cite{simonyan2014deep}, later work extended this method by overriding the backwards ReLU operator~\cite{zeiler2014visualizing,springenberg2015striving}. As these explanation maps are often noisy, methods such as \textsc{smoothgrad}~\cite{smilkov2017smoothgrad} and Integrated Gradients~\cite{sundararajan2017axiomatic} (IG) have been introduced to make the explanations more class specific.

\textbf{CAM-based methods} (Class Activation Map methods) were first introduced in \cite{zhou2016learning} and use a weighted sum of the feature maps generated by the last convolutional layer in a network as explanations. In the original paper, the weights were calculated using a Global Average Pooling layer, however later work explored different avenues to generate weights. Several later publications \cite{selvaraju2017grad,chattopadhay2018grad,omeiza2019smooth,fu2020axiom} use the gradients of the feature maps as weights, with some minor variations, while others follow a gradient-free approach due to issues such as gradient saturation. Examples of the second kind include Score-CAM ~\cite{wang2020score} which uses the \emph{channel-wise Increase In Confidence score}, and Ablation-CAM~\cite{ramaswamy2020ablation} which uses an equation to calculate a discrete gradient.

\textbf{Backpropagation-based methods} use the backpropagation mechanism to distribute the prediction score over the neural network nodes back to the pixel-level. This category includes algorithms such as Deep Taylor Decomposition~\cite{montavon2017explaining}, Layerwise Relevance Propagation (LRP)~\cite{bach2015pixel}, DeepLIFT~\cite{shrikumar2017learning} and Excitation Backpropagation~\cite{zhang2018top}. 

\subsection{Evaluating explanation methods}
\label{sec:evaluation-methods}

In the literature, there are four distinct categories of evaluation methods for visual explanations : Proxy tasks, ground truth-centered, model-centered, and human-centered evaluation.

\textbf{Proxy tasks} quantify the effectiveness of explanation methods by their performance in other tasks such as weakly supervised object localisation and weakly supervised object segmentation. Methods that have been introduced are the pointing game~\cite{zhang2018top} and outside-inside relevance ratio~\cite{lapuschkin2016analyzing}. These methods follow an assumption that a high performance in a proxy task translates to a high performing explanation.

\textbf{Ground truth-centered} methods depend on datasets that have a ground truth mask, i.e., pixel-wise annotations for each image, that indicate the relevant pixels for different labels. The annotations are typically dependent on the human notion of relevance and as such there can be a disconnect between annotated (human-relevant) pixels and the actual model-relevant pixels. This might lead to explanations seemingly failing in an unintuitive way, such as in the case of the Clever Hans phenomenon~\cite{lapuschkin2019unmasking}. To combat this, artificial datasets are used where the researcher has absolute control over the relevant features. These datasets come with the drawback that they are typically much less complex than standard datasets depicting more real-world settings. This category includes methods such as the pointing game~\cite{zhang2018top}, Relevant Feature Coverage~\cite{oramas2018visual}, Relevance Mass~/~Relevance Rank~\cite{arras2022clevr} and ROC curves~\cite{tjoa2020quantifying}.

\textbf{Model-centered} methods determine the quality of a saliency map by masking out pixels from the input samples following a distribution based on the saliency map. By measuring the change in prediction score, these methods determine how important the pixels, as indicated by the heatmap, are for the prediction score. 
Commonly used methods are Average drop~\cite{chattopadhay2018grad}, deletion and insertion curves~\cite{petsiuk2018rise} and AOPC values~\cite{samek2016evaluating}.

\textbf{Human-centered (qualitative)} methods are user studies that compare the visual quality of saliency maps generated from different methods. A set of users is asked to grade different aspects of the saliency maps such as localization, discrimination and more. Examples of this type of methods in action are in \cite{ribeiro2016should,selvaraju2017grad}.

\subsection{Sanity Checks and Axioms}

Several efforts have tried to formalise the process of explanation, and provide a rigorous framework for explanations to adhere to, via sanity checks and axioms. \cite{sundararajan2017axiomatic} introduces the axioms of sensitivity and implementation invariance. \cite{nie2018theoretical} discover that both guided back-propagation and DeConvNet are doing (partial) image recovery instead of actual explanations and determine the root cause to be the modifications to the backward ReLU operator. Independently, \cite{adebayo2018sanity} introduces several parameter randomization tests as sanity checks for visual explanations. \cite{kindermans2019reliability} introduces the axiom of input invariance and demonstrate that they can fool several saliency methods by applying a transformation on the input such as a constant shift. Finally, recent work by \cite{tomsett2020sanity} introduced a number of sanity checks on the evaluation metrics for saliency methods.

\subsection{Related Studies}

\cite{tjoa2020quantifying} used an artificial dataset to study a set of visual explanation methods using recall, precision and ROC curves that measure how well the explanations align with the class-relevant features. \cite{arras2022clevr} evaluates visual explanation methods on a synthetic dataset using the VQA task, by asking questions for an image and determining whether the explanation highlights the correct object in the image. \cite{samek2016evaluating} conducted a study using a small set of explanation methods on three datasets consisting of real-world photos. \cite{yang2019benchmarking} proposed an evaluation protocol using a carefully selected dataset and a set of metrics to evaluate explanations based on relative feature attribution.

Additionally, many works that introduce their own method compare the results against a set of related methods. These are typically more limited tests and serve to illustrate their advancements. Example of these quantitative experiments are \cite{petsiuk2018rise, zhang2018top}.

\begin{figure*}
\centering
\begin{subfigure}{\linewidth}
\centering
\includegraphics[width=0.1\textwidth]{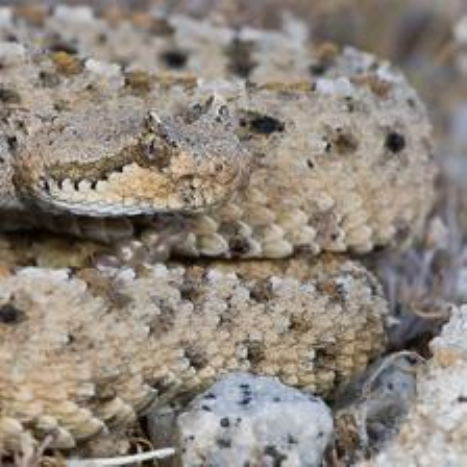}
\hspace{1em}
\includegraphics[width=0.1\textwidth]{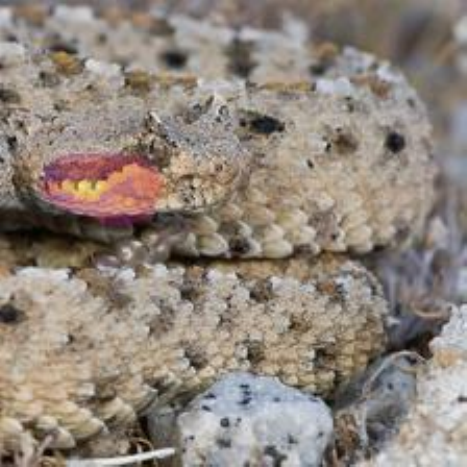}
\includegraphics[width=0.1\textwidth]{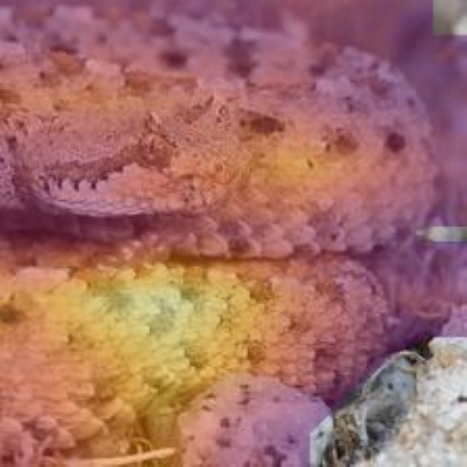}
\includegraphics[width=0.1\textwidth]{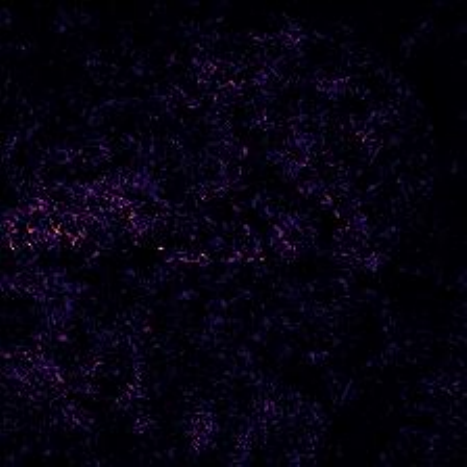}
\includegraphics[width=0.1\textwidth]{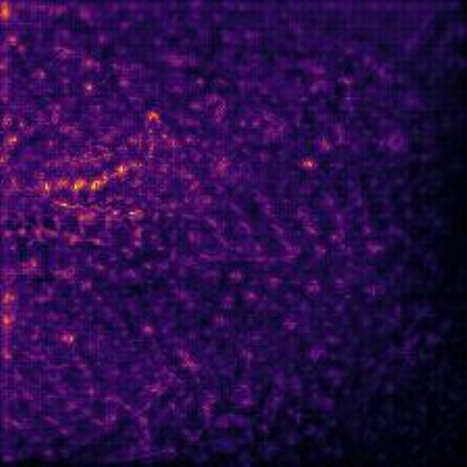}
\includegraphics[width=0.1\textwidth]{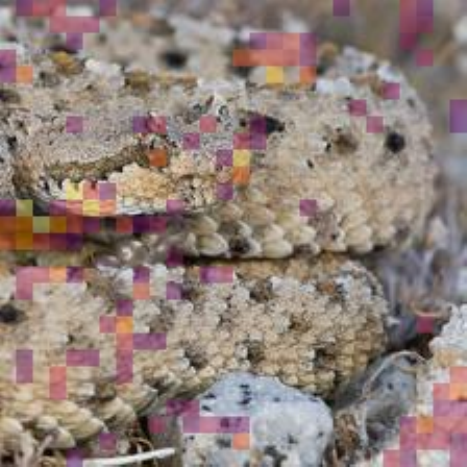}
\includegraphics[width=0.1\textwidth]{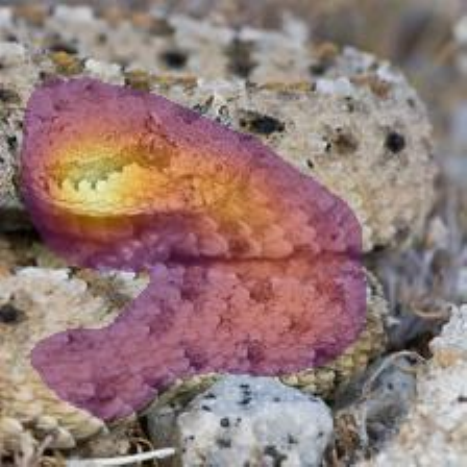}
\includegraphics[width=0.1\textwidth]{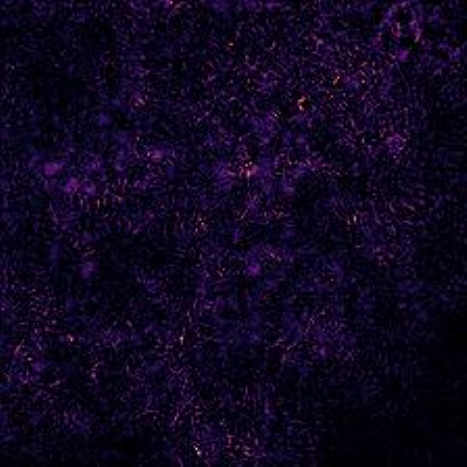}
\includegraphics[width=0.1\textwidth]{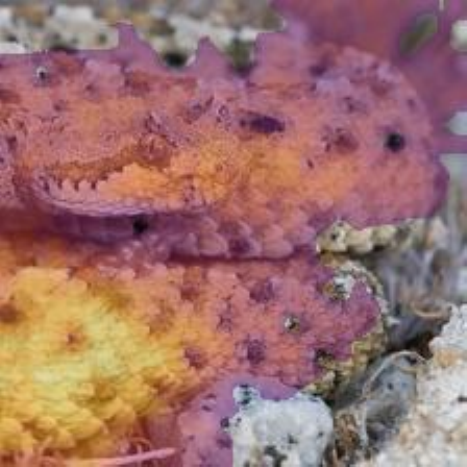}
\end{subfigure}
\begin{subfigure}{\linewidth}
\centering
\includegraphics[width=0.1\textwidth]{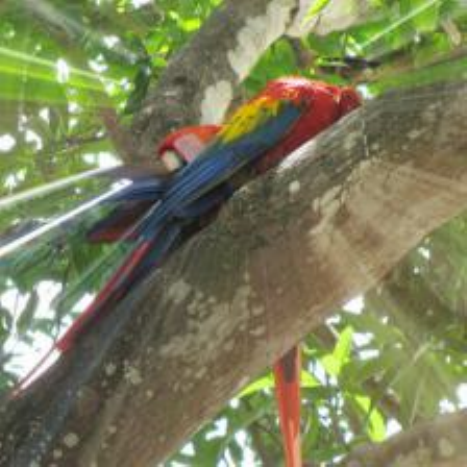}
\hspace{1em}
\includegraphics[width=0.1\textwidth]{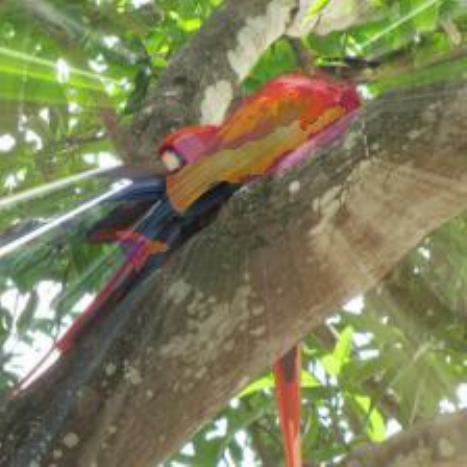}
\includegraphics[width=0.1\textwidth]{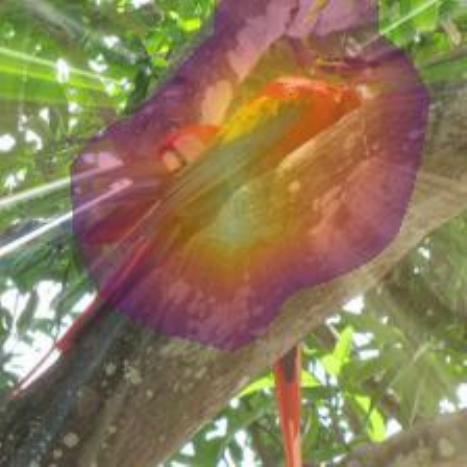}
\includegraphics[width=0.1\textwidth]{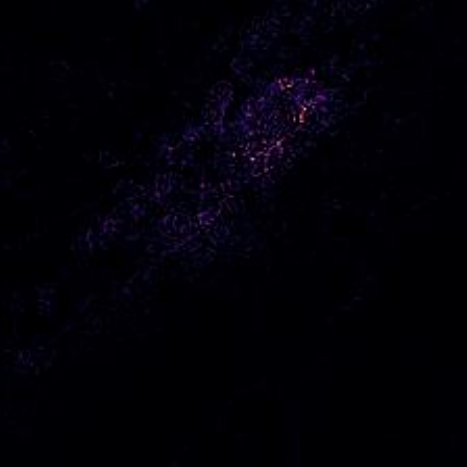}
\includegraphics[width=0.1\textwidth]{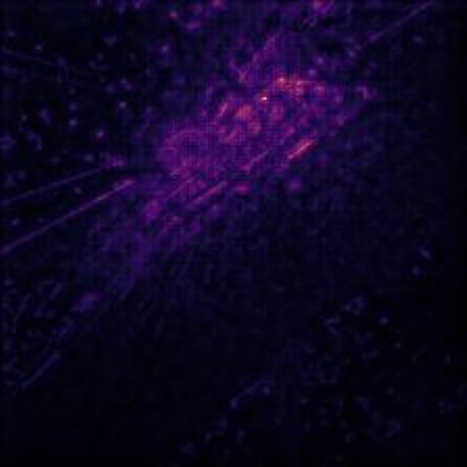}
\includegraphics[width=0.1\textwidth]{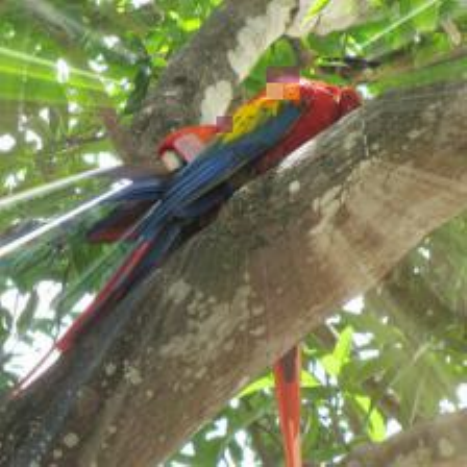}
\includegraphics[width=0.1\textwidth]{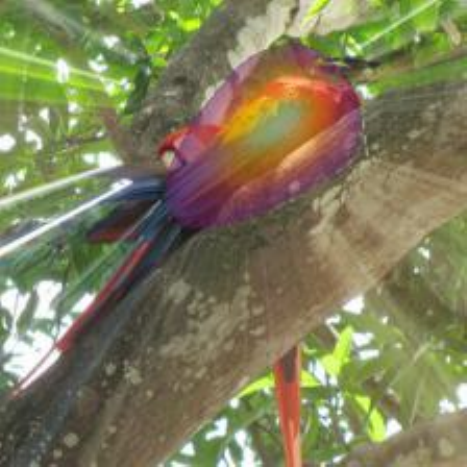}
\includegraphics[width=0.1\textwidth]{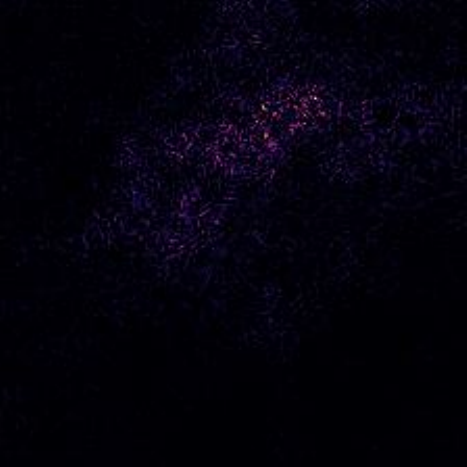}
\includegraphics[width=0.1\textwidth]{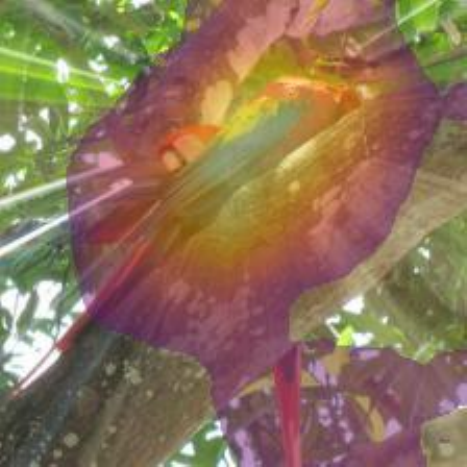}
\end{subfigure}
\begin{subfigure}{\linewidth}
\centering
\includegraphics[width=0.1\textwidth]{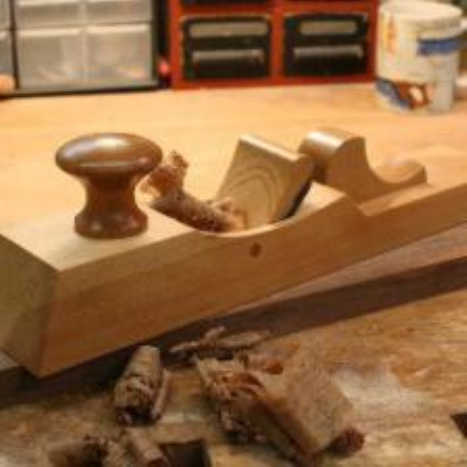}
\hspace{1em}
\includegraphics[width=0.1\textwidth]{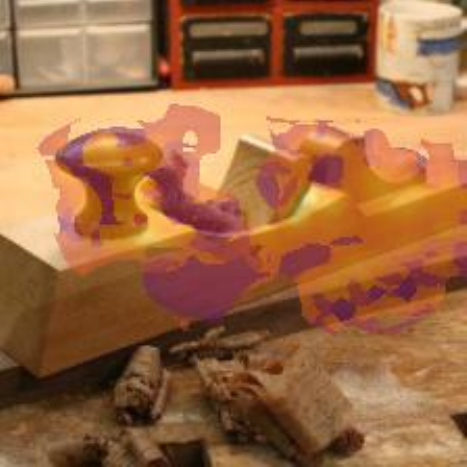}
\includegraphics[width=0.1\textwidth]{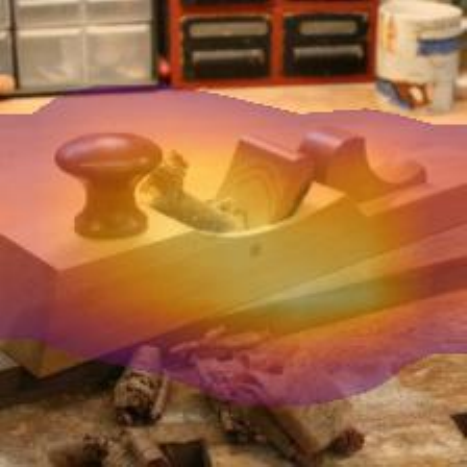}
\includegraphics[width=0.1\textwidth]{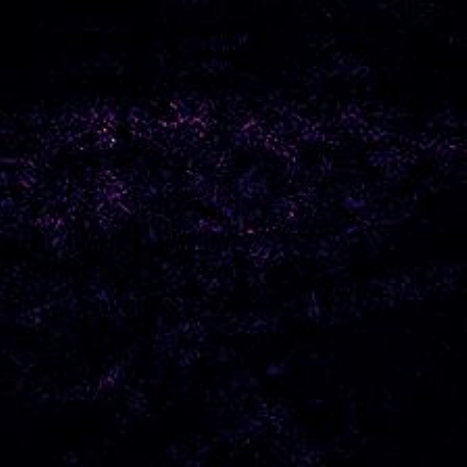}
\includegraphics[width=0.1\textwidth]{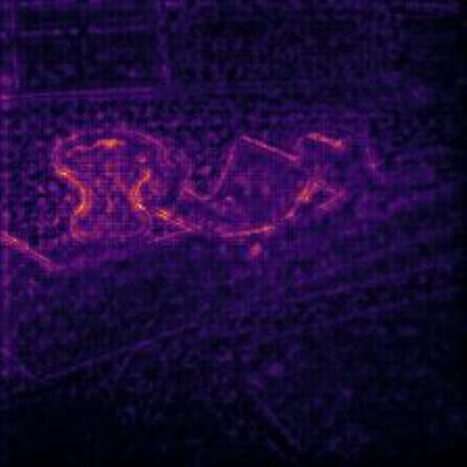}
\includegraphics[width=0.1\textwidth]{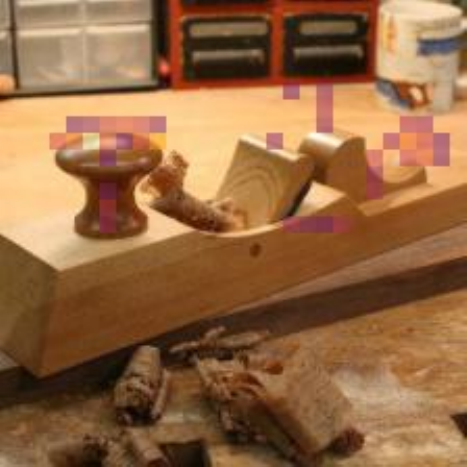}
\includegraphics[width=0.1\textwidth]{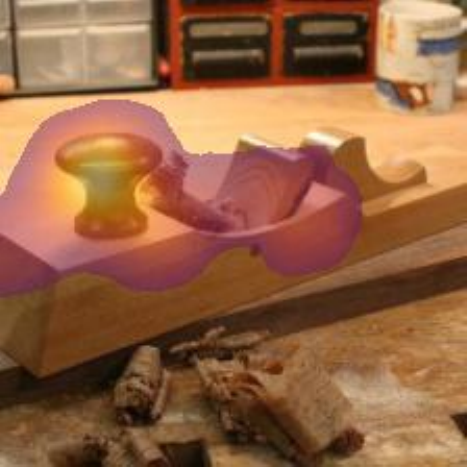}
\includegraphics[width=0.1\textwidth]{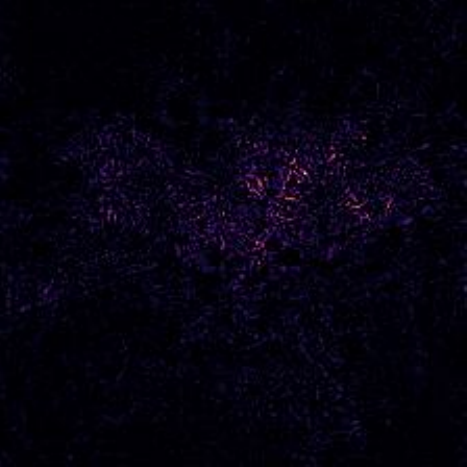}
\includegraphics[width=0.1\textwidth]{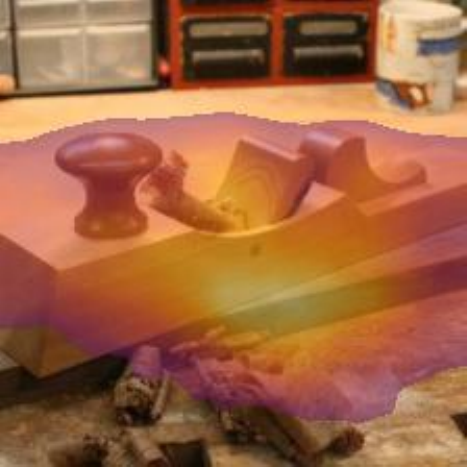}
\end{subfigure}
\begin{subfigure}{\linewidth}
\centering
\includegraphics[width=0.1\textwidth]{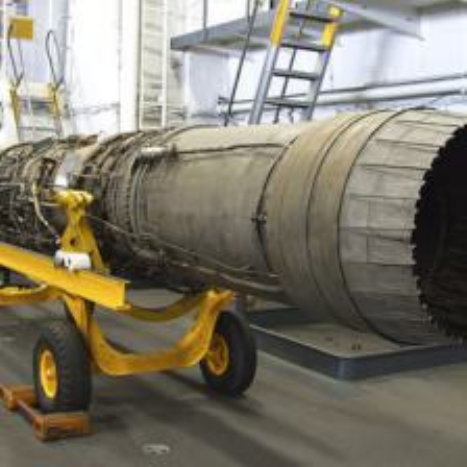}
\hspace{1em}
\includegraphics[width=0.1\textwidth]{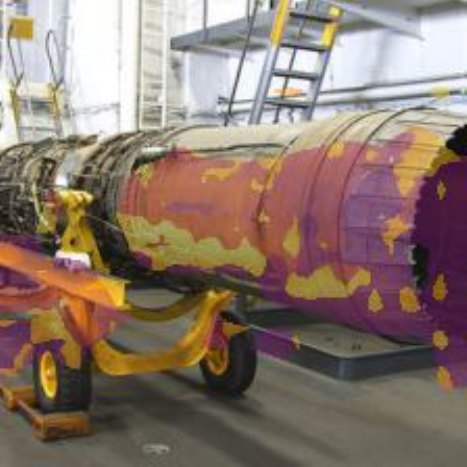}
\includegraphics[width=0.1\textwidth]{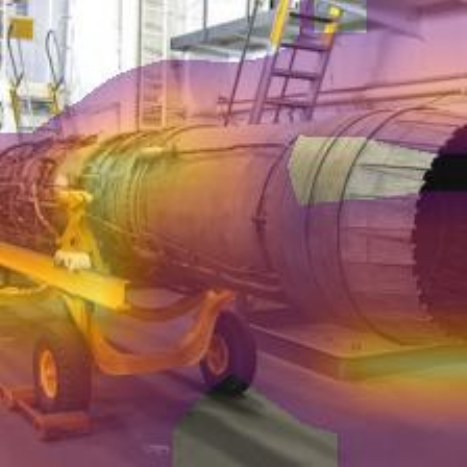}
\includegraphics[width=0.1\textwidth]{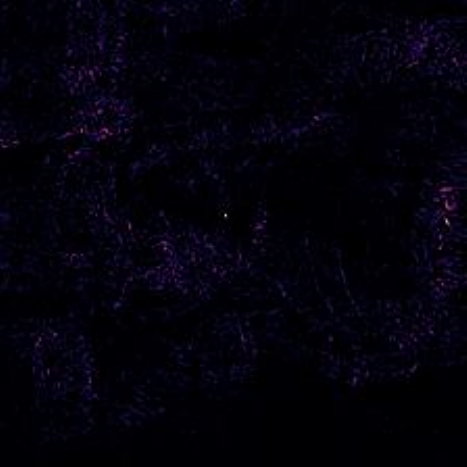}
\includegraphics[width=0.1\textwidth]{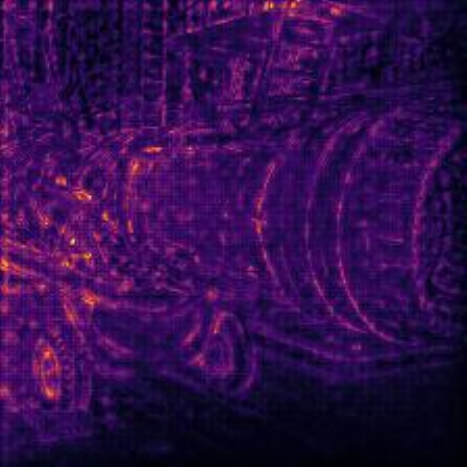}
\includegraphics[width=0.1\textwidth]{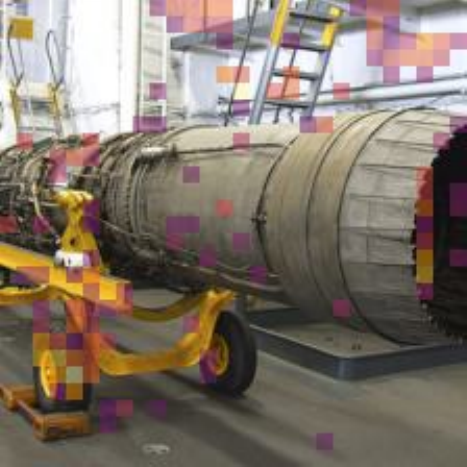}
\includegraphics[width=0.1\textwidth]{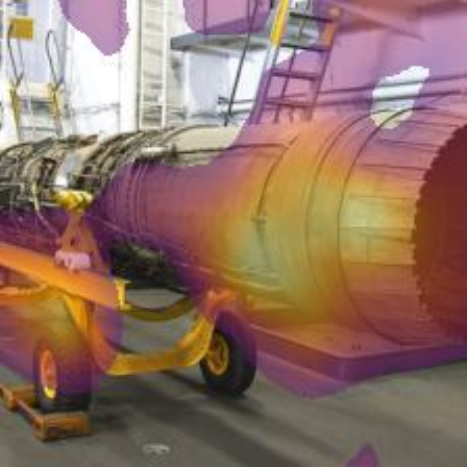}
\includegraphics[width=0.1\textwidth]{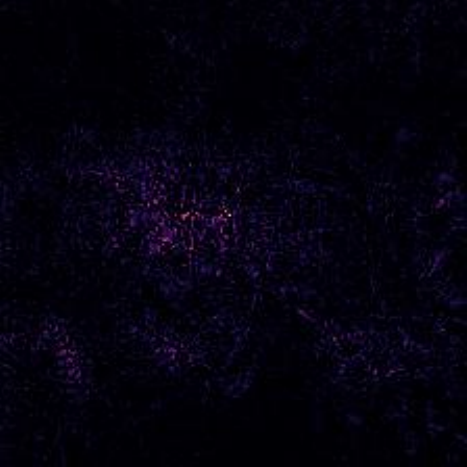}
\includegraphics[width=0.1\textwidth]{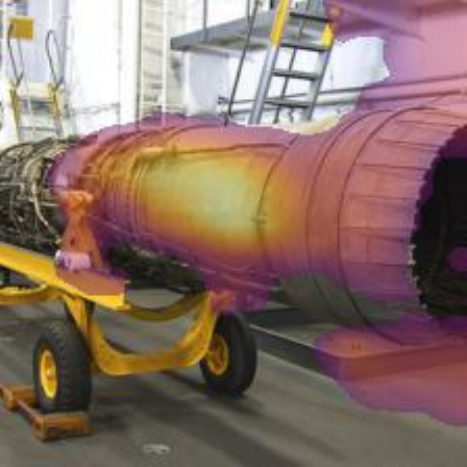}
\end{subfigure}
\begin{subfigure}{\linewidth}
\centering
\includegraphics[width=0.1\textwidth]{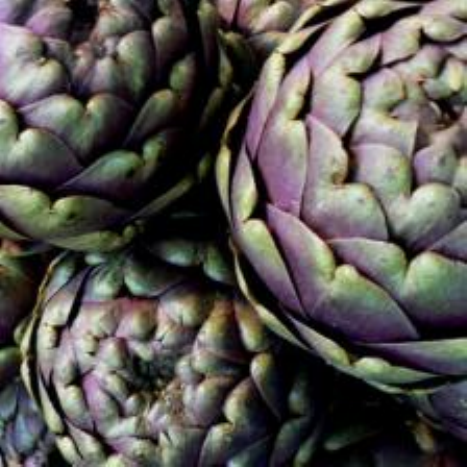}
\hspace{1em}
\includegraphics[width=0.1\textwidth]{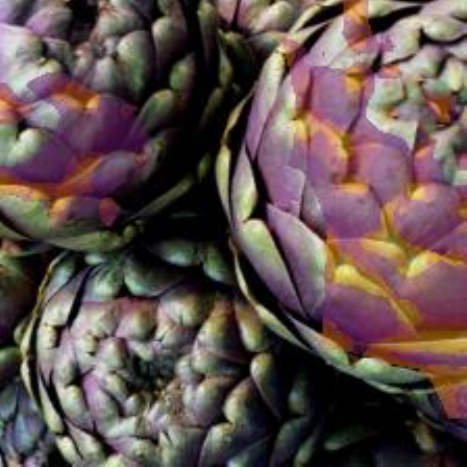}
\includegraphics[width=0.1\textwidth]{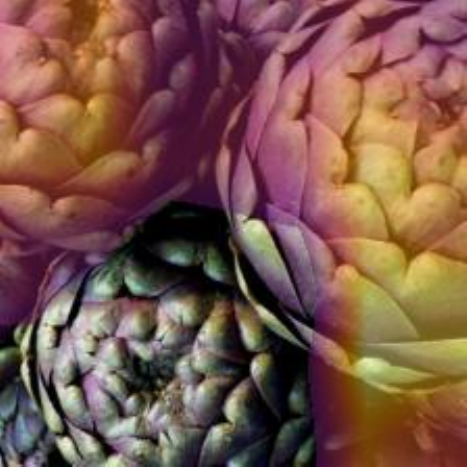}
\includegraphics[width=0.1\textwidth]{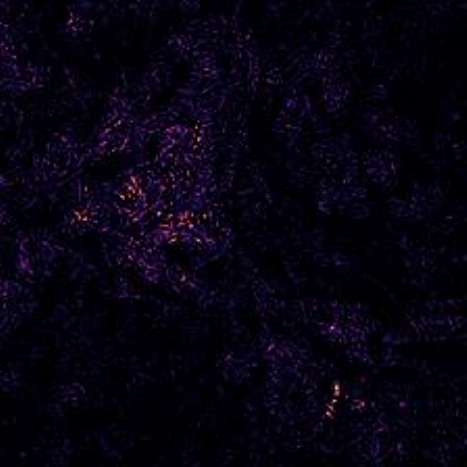}
\includegraphics[width=0.1\textwidth]{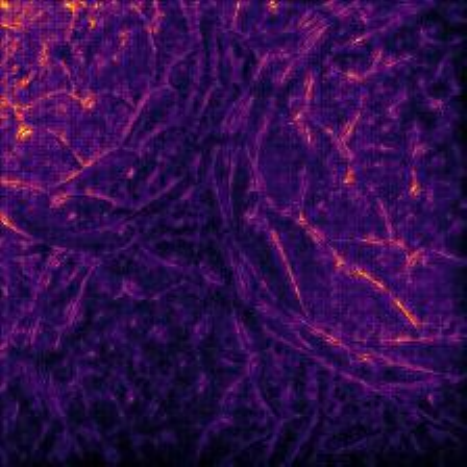}
\includegraphics[width=0.1\textwidth]{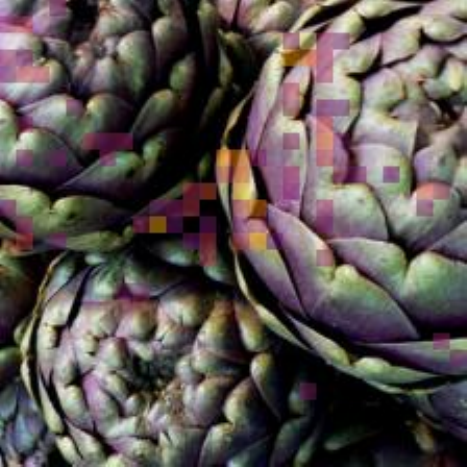}
\includegraphics[width=0.1\textwidth]{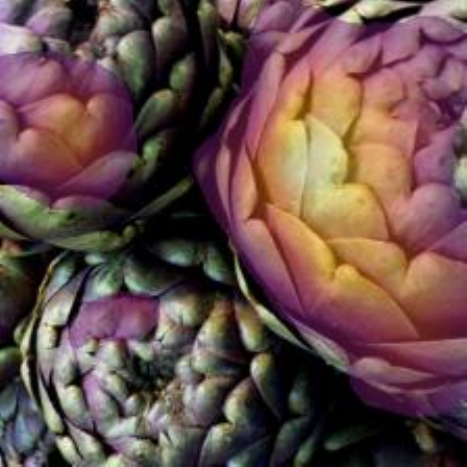}
\includegraphics[width=0.1\textwidth]{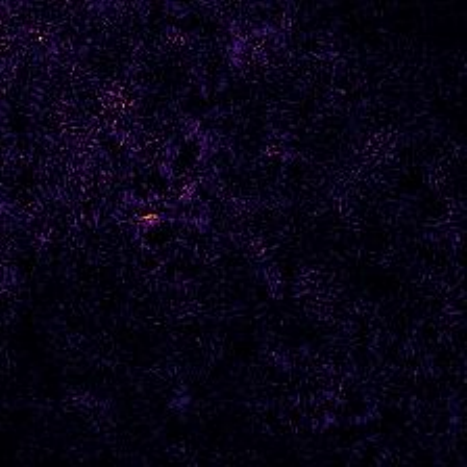}
\includegraphics[width=0.1\textwidth]{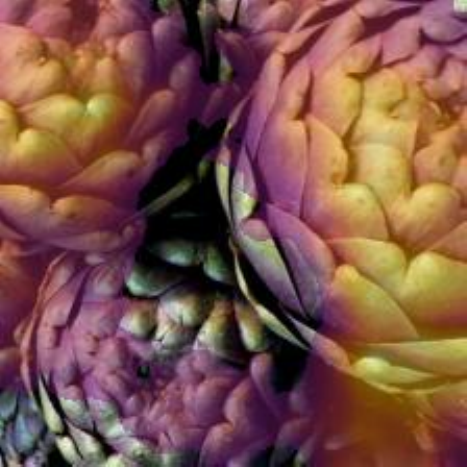}
\end{subfigure}
\vspace{1ex}
\centering
\includegraphics[width=0.3\linewidth]{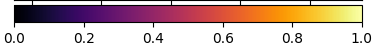}
\caption{Visual explanations for different samples using the ResNet-50 network. Also pictured is the color scale used to visualize the explanations. From left to right: input image, adaSISE, Grad-CAM, Integrated Gradients, LRP, occlusion, RISE, \textsc{smoothgrad}, TAME.}
\label{fig:resnet_demo}
\end{figure*}
\section{Methodology}
\label{sec:methodology}
This section will introduce the different components considered in our analysis. Moreover, it will provide an insight on the choice-decisions of such components.

\subsection{Datasets and Model Architectures}

\textbf{Datasets.} We use in our study the ImageNet 2012 validation set~\cite{deng2009imagenet, russakovsky2015imagenet}. This dataset is the de-facto baseline for Visual Explanation research~\cite{petsiuk2018rise, chattopadhay2018grad} and contains 1000 image classes with 50 samples per class.

\textbf{Models.} The architectures used in this paper are the VGG16 network~\cite{simonyan2014very} and ResNet-50~\cite{he2016deep}. To ensure reproducibility, we use the pre-trained weights for both networks provided by TorchVision~\cite{marcel2010torchvision}.

\subsection{Explanation Methods}

We focus our study on a subset of explanation methods (see \autoref{fig:resnet_demo} for sample visualisations). These methods were chosen to be representative of the breadth of the field while at the same time being sufficiently distinct from one another. In the paragraphs below, we describe the explanation methods we studied.

\textbf{Gradient-based methods} such as \textsc{smoothgrad}~\cite{smilkov2017smoothgrad} and Integrated gradients~\cite{sundararajan2017axiomatic} (IG) use the gradient $\partial F(x)_c/ \partial x$ as a measure of relevance for each pixel. However, the gradient is often noisy and discontinuous which can lead to artifacts. To remedy this, \textsc{smoothgrad} uses multiple slightly perturbed images and averages their gradients, such that only the most robust signals remains. On the other hand, IG uses a discrete integration technique by interpolating between a chosen uninformative baseline and the original image to generate a more robust gradient.

\textbf{CAM-based methods} such as Grad-CAM~\cite{selvaraju2017grad} use the fact that convolutional layers function as feature detectors and calculate a weighted sum of the feature maps from those convolutional layers for explanation. In this paper we only discuss Grad-CAM, which uses the mean of the gradients of the convolutional filter as weights, as it is the most widely used CAM-based method.

\textbf{Backpropagation-based methods} such as Layer-wise Relevance Propagation (LRP)~\cite{bach2015pixel} backpropagate a relevance score to the pixel level using different rules that determine how relevance is distributed between layers.

\textbf{Perturbation-based methods} such as (Sliding window) occlusion~\cite{zeiler2014visualizing} and RISE~\cite{petsiuk2018rise} generate explanations by perturbing parts of the input image with uninformative pixels and calculating the difference in prediction score of the perturbed image and the original image. This score difference is then used as a measure of importance of the perturbed part. To generate an explanation, an image is perturbed for a large number of iterations such that each pixel in the image is perturbed at least once. Occlusion accomplishes this by using a sliding window, while RISE uses a large amount of upsampled random masks generated from a Poisson distribution.

A special case is SISE~\cite{sattarzadeh2021explaining}, as this method combines aspects of CAM-based methods, backpropagation-based methods and perturbation-based methods. SISE can be classified as a variant of RISE, but rather than using random masks, it uses upsampled feature maps from within the model as masks. The feature maps that are used are selected based on the mean gradient for the feature map. In this paper, we consider an extension called AdaSISE \cite{sudhakar2021ada}, which adaptively sets the threshold on which feature maps are used as masks.

\textbf{Attention-based methods} such as TAME~\cite{ntrougkas2022tame} train an attention mechanism on top of the feature maps generated by a classification model to accurately generate explanations maps. In the case of TAME, this is enabled by a custom loss function that combines Cross Entropy, an area-based loss, and a variation loss to generate high-quality explanations.

\subsection{Evaluation Methods}

From the categories discussed in \autoref{sec:evaluation-methods}, we will limit ourselves and set the focus of our study on the \emph{model-centered evaluation methods} and the \emph{evaluations by proxy task}. We will motivate this briefly.
Applying a ground truth-centered evaluation is impossible as the used dataset (see \autoref{sec:methodology}) has no pixel-level ground truth annotations. An alternative to this would be to adopt a dataset related to semantic segmentation tasks~\cite{yu2018methods}. However, this is not necessarily correct given that the pixel-level annotations from segmentation tasks do not necessarily correspond to valid explanations. In few words, accurate object localization does not imply accurate explanation.
Finally, since we focus our study on a quantitative evaluation of the considered methods, we avoid the use of qualitative evaluation protocols, as these mostly focus on the subjective side of the analysis.
%


\textbf{The Average Drop \% score and Increase-In-Confidence score}~\cite{chattopadhay2018grad} is calculated by performing an element-wise multiplication of the saliency map $\mathcal{S}_{c}$ for class c with the input image $\mathcal{I}$. This results in an image where the pixel intensities are proportional to the relevance in the saliency map. This modified image is then used as input for the neural network model $F$, resulting in a modified prediction score. The average drop \% metric is then calculated by

\begin{equation}
\textrm{avg. Drop} = \frac{max(0, F(\mathcal{I})_c - F(\mathcal{S} \circ \mathcal{I})_c)}{F(\mathcal{I})_c}
\end{equation}
while increase-in-confidence is defined by

\begin{equation}
    \textrm{Confidence Increase} = F(\mathcal{S} \circ \mathcal{I})_c > F(\mathcal{I}_c)
\end{equation}
The intuition behind this metric is that
a good explanation heatmap should give higher values to relevant parts of the input. Therefore, no significant difference in the prediction score is to be expected if only those pixels are preserved. On the other hand, if the prediction score increases (Increase-in-Confidence), this means that conflicting information that was present within the original image, is no longer present in the modified image, thus meaning that the explanation does not highlight the conflicting information.
%
%
A low Average Drop \% indicates that the explanation method correctly identified the most relevant pixels in the input.
%

\textbf{The Insertion and Deletion metrics}~\cite{petsiuk2018rise} are metrics that start with an uninformative baseline/the original image from which pixels are progressively inserted/replaced by uniform values. The pixels are inserted/deleted based on their saliency score, with higher scoring pixels being inserted/deleted first.
Following each step the prediction score is recalculated. This process is conducted gradually until all the pixels are inserted or deleted. Then, an insertion/deletion curve is produced by plotting the scores against the percentage of inserted/deleted pixels. 
For easier comparison, the curves are quantified by calculating the Area Under Curve (AUC).

Intuitively, if the AUC of the insertion curve is high, then the pixels that are inserted first in the image have a large impact on the prediction score, which indicates that the relevance assigned by the explanation method correlates well to the important pixels. An equivalent reasoning can be applied to deletion curves with a low AUC.

\textbf{The Remove-and-Debias metric}~\cite{rong2022consistent} is a modification of the deletion metric, stemming from an observation that it is possible to infer class-specific information purely by using the masks generated during the deletion steps. The authors solve this issue by introducing a minimally revealing imputation, called Noisy Linear Interpolation. Rather than replacing the removed pixels by fixed values, the authors use the observation that neighbouring pixels are highly correlated, to generate replacements by solving a system of sparse linear equations based on the neighbouring pixel values.


\textbf{The Pointing Game metric}~\cite{zhang2018top} is a way to evaluate explanations using the proxy task of weakly supervised localisation. 
%
Under the Pointing game, an explanation is deemed good if its highest scored pixel lies inside the ground truth bounding box associated with the predicted label. This implies that good explanation method correctly localises objects related to the prediction. 
%
%
%
%

\begin{figure*}
\centering
\includegraphics[width=0.45\linewidth]{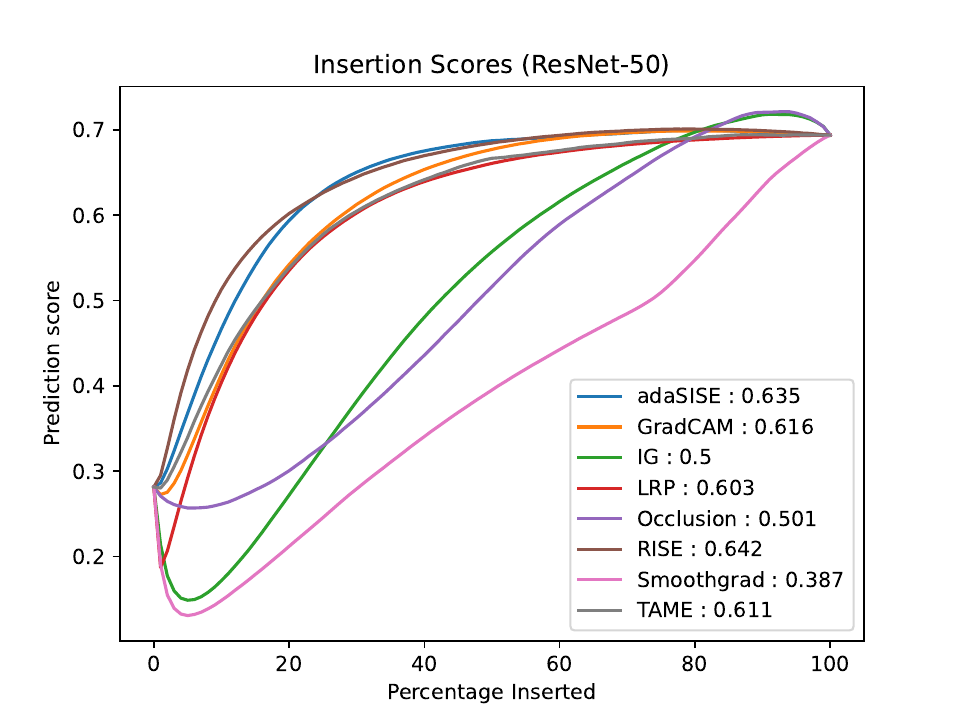}
\includegraphics[width=0.45\linewidth]{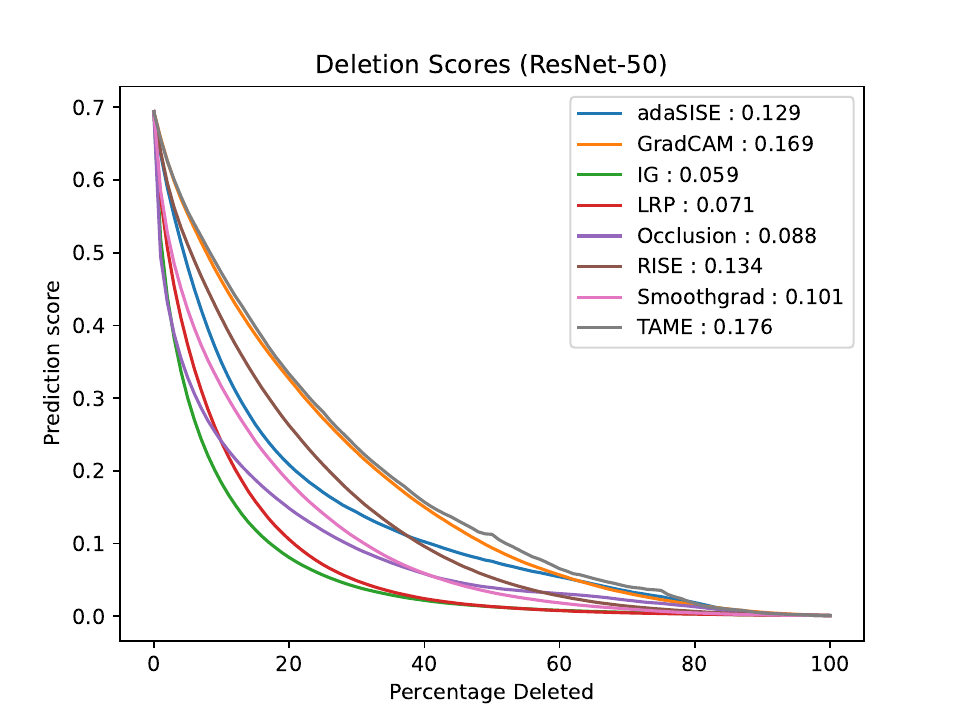}

\begin{tabular}{c|c|c|c|c|c|c|c|c}
& adaSISE & Grad-CAM  & IG & LRP & Occlusion & RISE & \textsc{smoothgrad} & TAME \\ \hline \hline 
Pointing Game ($\uparrow$) & \textbf{92.77\%} & 86.30\% & 81.72\% & 80.31\% & 83.66\% & 91.94\% & 89.45\% & 83.26\% \\ \hline
Avg. Drop \% ($\downarrow$) & 47.05\% & 14.96\% & 96.85\% & 66.26\% & 94.77\% & \textbf{14.02\%} & 96.99\% & 27.45\%\\ \hline
I.i.C. ($\uparrow$) & 20.04\% & 41.51\% & 1.81\% & 9.40\% & 2.68 \% & \textbf{43.90\%} & 1.67\% & 35.93\% \\ \hline
ROAD ($\downarrow$) & \textbf{0.166} & 0.201 & 0.292 & 0.199 & 0.238 & 0.170 & 0.447 & 0.206\\ \hline
\end{tabular}

\caption{The resulting scores when evaluating the considered explanation methods. Top: The insertion and deletion curves. Bottom : the other evaluation results in tabular form.}

\label{fig:resnet_insert}
\end{figure*}

\subsection{Sanity checks for evaluation metrics}
\label{sec:sanity_intro}

To complement our study, we will apply sanity checks proposed in the literature. These checks will allow us to assess the level of fidelity that the considered evaluation metrics have when applied on the generated explanations.

As in \cite{tomsett2020sanity}, we apply sanity checks on the studied evaluation metrics. In particular, we study the \emph{Internal consistency reliability} and \emph{Inter-method reliability}. 

\textbf{The Internal consistency reliability} is a measurement that determines how consistent the results from an evaluation metric are for different configurations of the metric. To determine the correlation between different configurations, we use the pairwise Spearman rank correlation.

\textbf{Inter-method reliability} indicates whether two metrics have the same definition of \emph{goodness} of an explanation. To determine whether two metrics follow a similar definition of 'goodness', we calculate a correlation score. In the case of two methods that produce continuous values, we use the Spearman rank correlation, while in the case that one method produces binary values (e.g. Pointing Game) and the other continuous values, we use the point-biserial correlation. Finally, When two metrics have opposing orderings (or ranks) --- i.e., for one metric higher values are better, while for the other metric lower values are better --- we first fix the ordering of both metrics to the same ordering. This is possible within the Spearman correlation, as it is a correlation of the ordering instead of the actual values. 

\subsection{Implementation Details}

Following common procedure, the images in the dataset are rescaled to a (224, 224) image and normalised using the ILSVRC-2012 mean ([0.485, 0.456, 0.406]) and standard deviation ([0.229, 0.224, 0.225]) before the explanation step.

The raw explanation maps are postprocessed using the following steps: First, negative relevance for the given label is removed. Then, the explanation map is min-max normalised to [0,1]. 
These steps serve to ease comparison and visualisation of explanation maps and follow common procedure, but introduce a pitfall, namely that care needs to be taken when comparing explanation maps. Due to the min-max normalisation, the relevances are now expressed in a relative factor instead of an absolute factor.


To improve the reproducibility of this study, our code and setup will be made publicly available\footnote{\scriptsize\url{https://github.com/Benjamin-Vandersmissen/quantitative-evaluation}}.

\section{Experiments}
\label{sec:experiments}

\subsection{Evaluating Explanations}
\label{sec:eval}

This first experiment will focus on quantitatively positioning
each of the considered explanation methods based on the evaluation metrics (\autoref{sec:evaluation-methods}). We do this by following the default configuration of each metric. Specifically for insertion and deletion this is pixel-level replacement with a dataset mean baseline (which is zero in this case). 
For easier comparison between the different configurations, we calculate the saliency map for the \emph{ground truth label} (regardless of predicted label) and evaluate the methods accordingly. This is necessary as the top-1 predictions by ResNet-50 and VGG16 are not always the same. In fact, on the considered dataset ResNet-50 achieves a top-1 classification accuracy of 75.6\%, while VGG16 achieves only 71.9\%.

When comparing results from different metrics, we can notice a number of trends that start to appear(\autoref{fig:resnet_insert}). Gradient-based methods together with LRP and Occlusion seem to perform worse on the insertion metric, the average drop metric and the Increase-in-Confidence, while performing better on the deletion metric and ROAD.
In contrast, the methods that produce a coarse saliency map (RISE, adaSISE, Grad-CAM and TAME) perform better on the insertion metric and the average drop metric, but worse on the deletion metric. Our hypothesis is that this occurs due to the sparsity of the generated explanations. We explore this in detail in \autoref{sec:sparsity}.
Finally, there seem to be no real connection between the pointing game scores and any other metrics, as the pointing game scores are similar for every method except LRP on the ResNet-50 network, which scores slightly lower. In that case, we can observe that several explanations generated by LRP have noise with high relevance around the edges of the images, which is an observation that doesn't hold true for the same method applied on VGG16.

\subsection{Internal Consistency Reliability}
\label{sec:internalConsistencyReliability}

In this experiment, we apply the \emph{Internal Consistency Reliability} (see \autoref{sec:sanity_intro}). In particular, we study the insertion and deletion metric as these metrics have a number of possible configurations introduced in the literature~\cite{samek2016evaluating, petsiuk2018rise}. 
The first of the two parameters we study is the \emph{uninformative value used}, which can be one of: the dataset mean (which is zero), a blurred version of the image or random uniform noise. In the case of blurring, a Gaussian kernel of size 11 with a $\sigma$ of 5 is used.
The second parameter is whether we insert/delete purely based on the highest relevance, or whether we insert/delete in a 9x9 neighbourhood around the highest relevant pixel --- similar to the AOPC values introduced in \cite{samek2016evaluating}. 

A quick glance at \autoref{tab:config_correl} clearly shows a high correlation between the different possible configurations for the insertion and deletion metric. It should be noted that blurring has a slightly lower correlation with the other uninformative values, which might be due to the fact that blurring the image removes less information as opposed to replacing by the mean or random noise. Due to space constraints, we limit ourselves to only showing four configurations each for insertion and deletion, namely the configurations originally introduced in the literature. A full set of configurations as well as accompanying visualisations can be found in the supplementary material.

\begin{table*}
\centering
\caption{The average pairwise Spearman rank correlation between different configurations of the insertion and deletion metric, calculated for ResNet-50.}
\begin{tabular}{c|c|c|c|c}
\textbf{Insertion} & mean+pixel & blur+pixel & mean+region & blur+region \\ \hline \hline 
mean+pixel & / & 0.816 & 0.873 & 0.775 \\ \hline 
blur+pixel & 0.816 & / & 0.786 & 0.948 \\ \hline 
mean+region & 0.873 & 0.786 & / & 0.807 \\ \hline 
blur+region & 0.775 & 0.948 & 0.807 & / \\  

\textbf{Deletion} & mean+pixel & random+pixel & mean+region & random+region \\ \hline \hline 
mean+pixel & / & 0.965 & 0.890 & 0.881 \\ \hline 
random+pixel & 0.965 & / & 0.872 & 0.897 \\ \hline 
mean+region & 0.890 & 0.872 & / & 0.966 \\ \hline 
random+region & 0.881 & 0.897 & 0.966 & / \\  

\end{tabular}
\label{tab:config_correl}
\end{table*}

\subsection{Inter-method reliability}
\label{sec:inter-methodReliability}
Now we focus our study on the other sanity check, \emph{Inter-method Reliability} (\autoref{sec:sanity_intro}). This time we reuse the evaluation results gotten from the first experiment to calculate the pairwise reliability between every pair of metrics. We report the computed pairwise correlation values in \autoref{tab:correlation}.

We can observe several trends in the computed correlation results (see \autoref{tab:correlation}).
First, the insertion and deletion metrics have a high negative correlation, meaning that explanations that have a higher insertion score, typically have a lower deletion score. This suggests that the insertion and deletion metric have opposing notions of what a good explanation is. We will give an intuition to this observation in \autoref{sec:discussion}.
Second, the average drop \% metric has a relatively low correlation with the insertion and deletion metric, but exhibits and interesting behaviour. We can see that often the sign of the correlation is different between LRP, Grad-CAM and RISE on one hand and IG, Occlusion and \textsc{Smoothgrad} on the other hand. We hypothesise that this is related to the specific definition of average drop \% and elaborate further on this in \autoref{sec:avgdropexperiment}.
Third, the pointing game metric has very low correlation scores with any of the other metrics. This might indicate that the pointing game --- and by extension the weakly supervised object localization task --- is not a good proxy to measure the quality of visual explanations. This further support the observation that good localization does not imply good explanation.


\begin{table*}

\centering
\caption{The pairwise Spearman correlation (and point-biserial correlation) coefficient between different evaluation metrics, calculated for ResNet-50.}
\begin{tabular}{c|c|c|c|c|c|c|c|c}
Correlation coefficients & AdaSISE & Grad-CAM & IG & LRP & Occlusion & RISE & \textsc{smoothgrad} & TAME \\ \hline \hline
insertion \& deletion  & -0.662 & -0.695 & -0.677 & -0.659 & -0.644 & -0.676 & -0.715 & -0.64 \\ \hline 
insertion \& avg. drop  & 0.09 & 0.172 & -0.349 & 0.101 & -0.322 & 0.017 & -0.349 & 0.243 \\ \hline 
insertion \& pointing  & -0.179 & -0.137 & -0.135 & -0.155 & -0.129 & -0.212 & -0.131 & -0.215 \\ \hline 
insertion \& ROAD  & -0.715 & -0.724 & -0.788 & -0.698 & -0.746 & -0.712 & -0.9 & -0.668 \\ \hline 
deletion \& avg. drop  & 0.008 & -0.063 & 0.318 & -0.146 & 0.309 & 0.106 & 0.272 & 0.042 \\ \hline 
deletion \& pointing  & 0.060 & 0.084 & 0.014 & 0.040 & -0.038 & 0.088 & 0.049 & -0.051 \\ \hline 
deletion \& ROAD  & 0.9 & 0.918 & 0.781 & 0.81 & 0.853 & 0.904 & 0.698 & 0.928 \\ \hline 
avg. drop \& pointing  & -0.036 & -0.047 & 0.076 & -0.056 & 0.070 & -0.089 & 0.084 & -0.3 \\ \hline 
avg. drop \& ROAD  & 0.075 & -0.049 & 0.405 & 0.011 & 0.369 & 0.127 & 0.398 & 0.061 \\ \hline 
pointing \& ROAD  & -0.062 & -0.080 & -0.086 & -0.047 & -0.000 & -0.084 & -0.129 & 0.059 \\ \hline 
\end{tabular}
\label{tab:correlation}
\end{table*}


\subsection{Explanation Sparsity}
\label{sec:sparsity}

\begin{figure}
\centering
\includegraphics[width=0.9\linewidth]{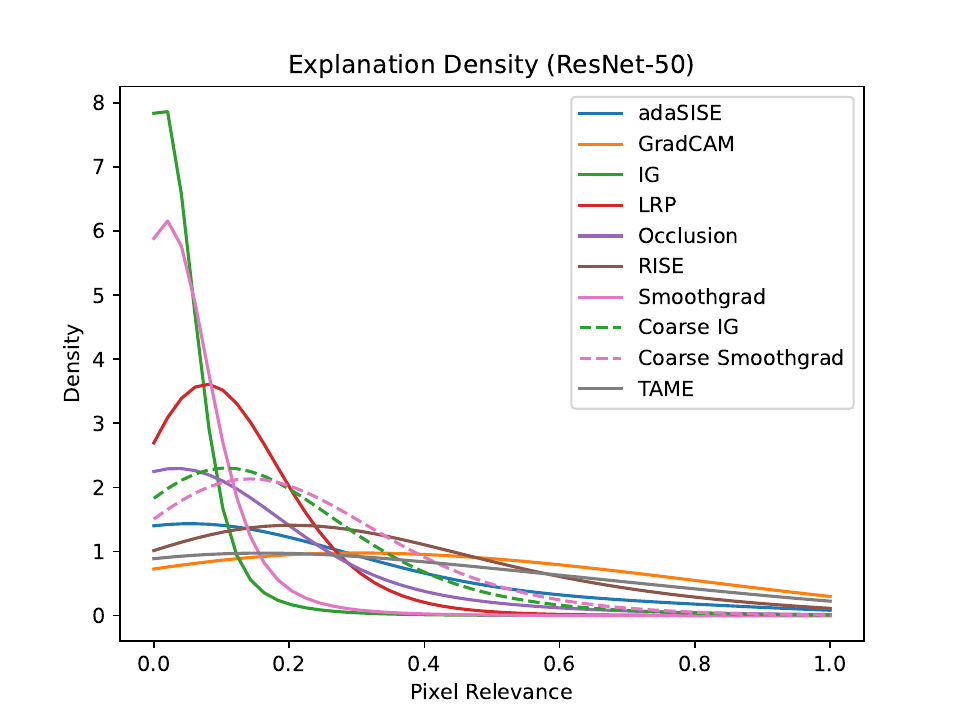}
\caption{The kernel density estimation of the relevance assigned by different explanation methods.}
\label{fig:sparsity}
\end{figure}

\begin{figure*}
\centering
\includegraphics[width=0.45\linewidth]{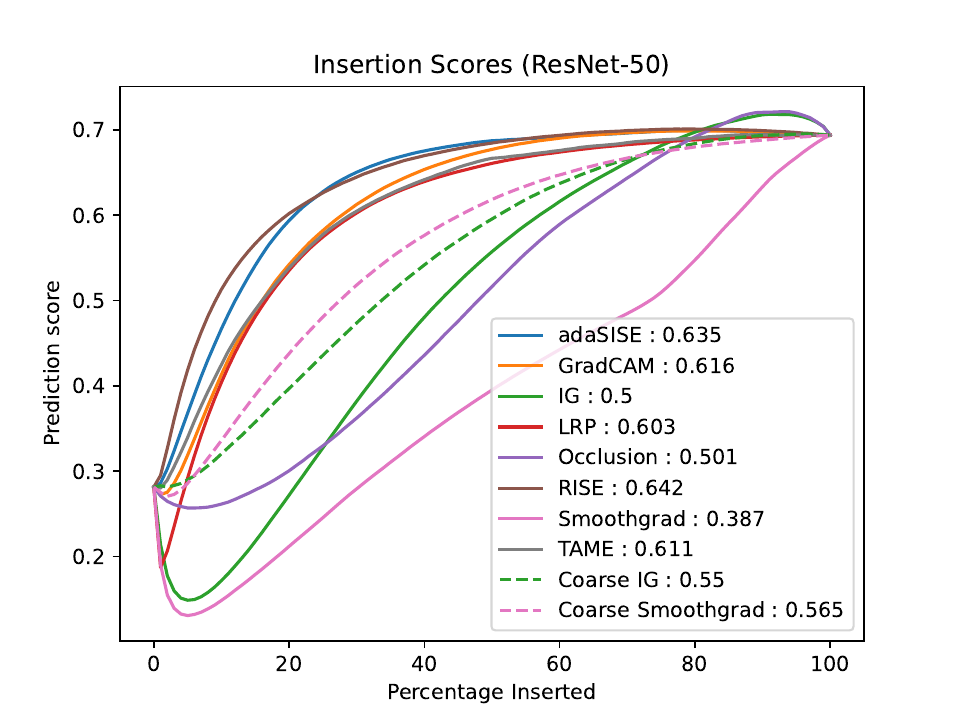}
\includegraphics[width=0.45\linewidth]{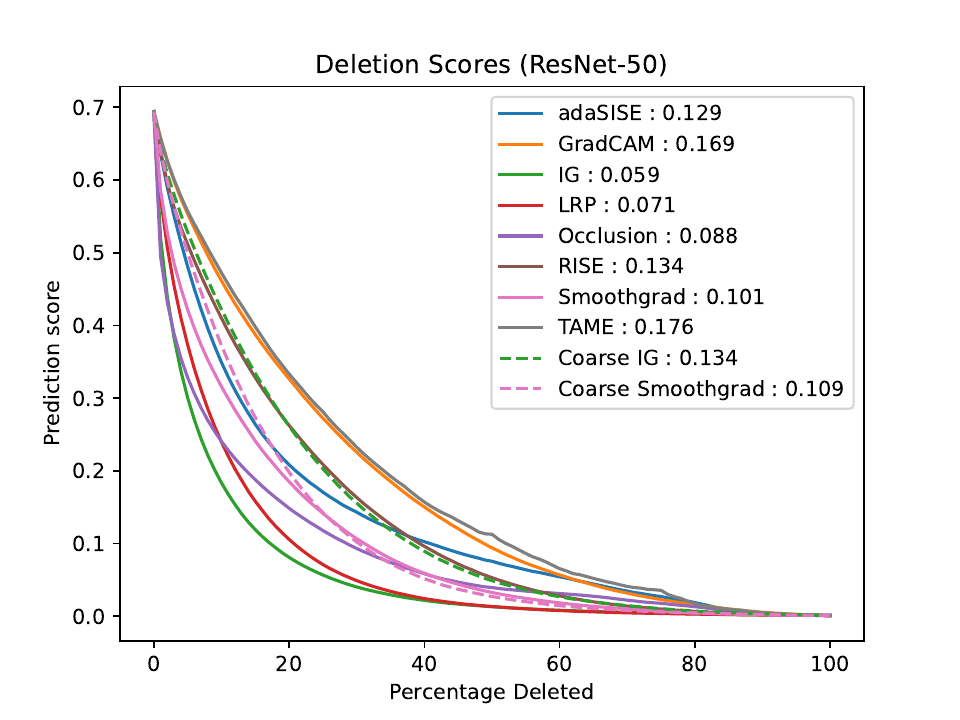}

\begin{tabular}{c|c|c|c|c|c|c|c|c}
\small
& adaSISE & Grad-CAM  & IG (\textit{*})& LRP & Occlusion & RISE & \textsc{smoothgrad} (\textit{*}) & TAME \\ \hline \hline 
Pointing ($\uparrow$) & 92.77\% & 86.30\% & 82.12\% {(\footnotesize81.72\%)}& 80.31\% & 83.66\% & \textbf{91.94\%} & 88.96\% {\footnotesize(89.45\%)} & 83.26\% \\ \hline
Drop \% ($\downarrow$) & 47.05\% & 14.96\% & 44.99\% {\footnotesize(96.85\%)} & 66.26\% & 94.77\% & \textbf{14.02\%} & 49.11\% {\footnotesize(96.99\%)} & 27.45\%\\ \hline
I.i.C. ($\uparrow$) & 20.04\% & 41.51\% & 21.88\% {\footnotesize(1.81\%)} & 9.40\% & 2.68 \% & \textbf{43.90\%} & 15.12\% {\footnotesize(1.67\%)} & 35.93\% \\ \hline
ROAD ($\downarrow$) & \textbf{0.166} & 0.201 & 0.213  {\footnotesize(0.292)} & 0.199 & 0.238 & 0.170 & 0.185 {\footnotesize(0.447)} & 0.206\\ \hline
\end{tabular}

\caption{Comparing blurred versions of IG and \textsc{smoothgrad} explanations to the other explanation methods. Top: The insertion and deletion curves. Bottom : the other evaluation results in tabular form (Original values of IG and \textsc{smoothgrad} are provided between brackets).}
\label{fig:insert_coarse_grads}
\end{figure*}

In our observations, gradient-based methods such as IG and \textsc{Smoothgrad} generate very sparse explanations, while methods such as RISE and Grad-CAM generate very coarse explanations. This is demonstrated in \autoref{fig:sparsity} by using a Gaussian Kernel Density Estimation over all explanations in the test set. These fundamental differences stem from the fact that gradients are by nature very sparse, while the explanations from Grad-CAM and RISE are coarse due to upsampling of very small feature maps.
To study whether this sparsity has a large impact on the evaluation scores of the Gradient-based methods, we apply a Gaussian blur on their explanation maps. The goal of this step is to artificially induce coarseness, similar as in \cite{zhang2018top}. We then repeat the same steps from the experiment in \autoref{sec:eval} and discuss the impact of the blurring step below with regard to the original explanations. See \autoref{fig:insert_coarse_grads} for the results of this experiment. 

It can be seen clearly that the resulting scores for the insertion, deletion and average drop \% metric are significantly impacted. On the one hand, we notice a significantly better insertion and average drop \% score and a slightly worse deletion score. On the other hand, there is only a negligible difference between the pointing game scores. 

The results regarding sparsity for the insertion/deletion metric are similar to the results we found when comparing the insertion/deletion by pixel with insertion/deletion by region experiment from \autoref{sec:internalConsistencyReliability}. In fact using the region-based insertion and deletion is a similar way to evaluate sparse explanations as coarse explanations. This because it perturbs pixels in a neighbourhood around a selected pixel, which typically have --- in the case of coarse explanations --- very similar values. Additional visualisations will be provided in the supplementary materials.

\subsection{Average Drop \% reformulated}
\label{sec:avgdropexperiment}
Here we further study the difference between average drop with real-valued explanations (as it is originally formulated) and average drop with binarized explanations on a per image level. We posit that using binary explanations are better for evaluation, as multiplication with a real-valued explanation can introduce unexpected effects. Examples are f.e. colour difference, which can lead to misclassification. 
We show the inter-method reliability between the binarized formulization and the other evaluation methods in \autoref{tab:binary_correl}. When comparing these results with the results from \autoref{sec:inter-methodReliability}, we notice that the behaviour with binarized average drop \% is much more inline with the expected behaviour. A more in-depth comparison with different levels of binarization, as well as the impact on the evaluation score can be found in the supplementary materials.

\begin{table}
    \centering
    \caption{Inter-method reliability between Binarized (50th percentile) Average Drop \% and other evaluation metrics.}
    \begin{tabular}{c|c|c|c|c}
          & Insertion & Deletion & Pointing & ROAD \\ \hline \hline
adaSISE & 0.1361 & -0.0344 & 0.00422 & 0.0256 \\
Grad-CAM & 0.2094 & -0.1615 & 0.0309 & -0.0793 \\
IG & 0.2411 & -0.2626 & 0.0278 & -0.0365 \\
LRP & 0.2398 & -0.2617 & 0.1528 & -0.1065 \\
Occlusion & 0.0169 & 0.0916 & 0.0013 & 0.1683 \\
RISE & 0.1328 & -0.0782 & 0.0838 & 0.0152 \\
\textsc{smoothgrad} & 0.2285 & 0.0227 & 0.0344 & -0.1306 \\
TAME & 0.2645 & -0.0282 & 0.2763 & 0.0251 \\
    \end{tabular}
    \label{tab:binary_correl}
\end{table}

\section{Discussion}
\label{sec:discussion}

\textbf{Evaluation metric configurations.}
In \autoref{sec:internalConsistencyReliability} we demonstrated that using different configurations of the same evaluation metric can have an impact on the scoring of  explanation maps, but has only a limited impact on the ranking of explanation maps across different configurations of the evaluation metric. As such, it is difficult to quantify the exact goodness of an explanation, but it is much easier to determine the better explanation from a set of explanations.

\textbf{Coherency between metrics.}
Three groups of evaluation metrics can be identified in our test set. \textbf{(1)} Metrics such as insertion and average drop, which assign value to the  set of pixels that influence the prediction most when considered in a vacuum. \textbf{(2)} Metrics such as deletion, which assign value to the set of pixels that when removed from an image influence the prediction most. Finally, \textbf{(3)} metrics that consider proxy tasks, e.g. object detection or segmentation. By considering the correlation results in \autoref{sec:inter-methodReliability} and \autoref{sec:avgdropexperiment}, we can conclude that \textbf{(1)} and \textbf{(2)} encode opposite notions of "goodness", while \textbf{(3)} is almost uncorrelated to both \textbf{(1)} and \textbf{(2)}. This reflects upon the fact that localization is not necessarily explanation.

\textbf{Sparsity in explanations.}
Due to the nature of convolution, filters will typically activate for a region of connected pixels on which the filter is trained. As such metrics of group \textbf{(1)} show better performance on coarse explanation maps (see \autoref{sec:eval}). For evaluation methods of category \textbf{(2)}, we can make a similar argument why they perform better on Gradient-based explanations, namely due to the fact that when pixels with the highest gradient are removed, the prediction score drops immensely. This property has been used in f.e. adversarial attacks.

Artificially adding coarseness to gradient-based explanations in \autoref{sec:sparsity} reinforced these theoretical observations, as we can see that the metrics from group \textbf{(1)} record a better score, while the metrics from group \textbf{(2)} record a worse score.

As far as the impact of sparse explanations on metrics of group \textbf{(3)}, no significant difference can be noted, further reinforcing the observation that these metrics are unsuitable to evaluate explanations.

\textbf{Recommended evaluation protocol.} Given the results of this research, we advise against using proxy tasks for evaluation, such as the Pointing Game, weakly-supervised object localization~\cite{zhou2016learning} or semantic segmentation~\cite{selvaraju2017grad}.
These type of methods operate under the assumption that good explanation implies good object localization which is not the case; think about visual explanations produced for a fine-grained classification problem.

Additionally, while ground-truth based evaluation \cite{oramas2018visual,yang2019benchmarking} is the most accurate way to measure the performance of visual explanations, generating a ground-truth labelled dataset is highly expensive, is often still dependent on human notions of relevancy. Consequently, these type of methods operate under the assumption that there is an alignment between features considered by humans and those considered by machines when making predictions for which there is evidence of the contrary (\cite{vqahat})
Moreover, these datasets produced for these ground-truth based evaluations are often simpler than natural image datasets.

Model-based evaluation is mainly based on perturbing the input images based on the information contained within the explanation. Unfortunately, this has as side effect that the model is often evaluated with modified images that are not anymore within the data distribution. This is an issue that ROAD tries to avoid by solving large sets of sparse linear equations. However, this makes it such that the runtime of the evaluation is much higher. Furthermore, our experiments show that ROAD has a very significant correlation with simpler methods such as deletion with a blurred image. As such, we recommend to consider the trade-off between speed and theoretical soundness to determine which option to use for evaluation.

\section{Conclusion}
\label{sec:conclusion}
In this paper we have studied a number of commonly used explanation methods and quantitatively evaluated them using a variety of evaluation metrics from the literature. 
We found that the evaluation results for the explanations are generally inconclusive - different saliency methods perform better or worse on different metrics. This in turn makes it difficult to rank explanations based on an ensemble of evaluation metrics. 

Furthermore, elaborating on these empirical results, we analysed the evaluation metrics and found that at best, there is little correlation between metrics and at worst some metrics were shown to be contradictory. 
We identified two main trends in evaluation metrics: either an explanation is judged on how well the highest-scoring features perform in a vacuum, or an explanation is judged on how bad the image without the highest-scoring features performs. We also noticed that the pointing game is generally a poor metric to evaluate explanations. 
Finally, we studied the impact of the sparsity of visual explanations and its relation to the two categories of evaluation methods we outlined.

\section*{Acknowledgments}
\begin{itemize}
    \item FWO Fundamental Project G0A4720N \textit{”Design and Interpret: A New Framework for Explainable Artificial Intelligence”}
\end{itemize}

\bibliographystyle{ieeetr}
\bibliography{refs}

\clearpage
\newcommand{\beginsupplement}{%
        \setcounter{table}{0}
        \renewcommand{\thetable}{A\arabic{table}}%
        \setcounter{figure}{0}
        \renewcommand{\thefigure}{A\arabic{figure}}%
        \setcounter{section}{0}
        \renewcommand{\thesection}{\Alph{section}}%
     }

\onecolumn
\section*{Supplementary Material}
\beginsupplement
\section{Overview}
This supplementary contains additional visualisations for the different experiments mentioned in the main paper, as well as additional secondary insights.

In \autoref{sec:internal}, we highlight the results for the \textit{Internal Reliability Consistency} test in greater detail. Related to this, we provide some visualisations of the different types of uninformative values used in conjunction the deletion procedure in \autoref{sec:uninformative}. Additional visualisations illustrating the effect of Gaussian Blur to remove sparsity in visual explanations are shown in \autoref{sec:sparseness}. The average drop \% with binarised explanations is discussed in \autoref{sec:drop}. Finally, results for the VGG16 network are listed in \autoref{sec:vgg16}.

\section{Internal Reliability Consistency}
\label{sec:internal}

\begin{table*}
    \centering
    \caption{The correlations between different configurations of the insertion / deletion metric for the ResNet-50 network, broken down by explanation method. (1) = mean+pixel, (2) = uniform+pixel, (3) = blur+pixel, (4) = mean+region, (5) = uniform+region, (6) = blur+region}
    \label{tab:consistency_resnet}
    \begin{tabular}{c|c|c|c|c|c|c|c|c}
    \textbf{insertion} & adaSISE & Grad-CAM & IG & LRP & Occlusion & RISE & \textsc{smoothgrad} & TAME\\ \hline \hline
    (1) vs (2) & 0.969 & 0.973 & 0.926 & 0.975 & 0.93 & 0.966 & 0.948 & 0.975 \\
    (1) vs (3) & 0.851 & 0.865 & 0.745 & 0.846 & 0.72 & 0.847 & 0.804 & 0.854 \\
    (1) vs (4) & 0.968 & 0.966 & 0.658 & 0.903 & 0.891 & 0.947 & 0.684 & 0.974 \\
    (1) vs (5) & 0.946 & 0.949 & 0.683 & 0.897 & 0.852 & 0.926 & 0.71 & 0.955 \\
    (1) vs (6) & 0.843 & 0.857 & 0.611 & 0.81 & 0.704 & 0.837 & 0.695 & 0.848 \\
    (2) vs (3) & 0.837 & 0.852 & 0.744 & 0.834 & 0.691 & 0.832 & 0.809 & 0.84 \\
    (2) vs (4) & 0.948 & 0.95 & 0.64 & 0.886 & 0.849 & 0.932 & 0.658 & 0.957 \\
    (2) vs (5) & 0.973 & 0.972 & 0.695 & 0.909 & 0.899 & 0.957 & 0.702 & 0.978 \\
    (2) vs (6) & 0.83 & 0.845 & 0.638 & 0.799 & 0.681 & 0.823 & 0.708 & 0.835 \\
    (3) vs (4) & 0.845 & 0.861 & 0.686 & 0.833 & 0.705 & 0.843 & 0.665 & 0.848 \\
    (3) vs (5) & 0.829 & 0.843 & 0.689 & 0.82 & 0.68 & 0.819 & 0.674 & 0.83 \\
    (3) vs (6) & 0.994 & 0.998 & 0.827 & 0.968 & 0.962 & 0.997 & 0.839 & 0.998 \\
    (4) vs (5) & 0.969 & 0.974 & 0.95 & 0.961 & 0.927 & 0.965 & 0.951 & 0.974 \\
    (4) vs (6) & 0.842 & 0.857 & 0.772 & 0.827 & 0.719 & 0.838 & 0.756 & 0.845 \\
    (5) vs (6) & 0.826 & 0.839 & 0.754 & 0.807 & 0.693 & 0.815 & 0.741 & 0.827 \\
    
    \end{tabular}
    \begin{tabular}{c|c|c|c|c|c|c|c|c}
    \textbf{deletion} & adaSISE & Grad-CAM & IG & LRP & Occlusion & RISE & \textsc{smoothgrad} & TAME\\ \hline \hline
    (1) vs (2) & 0.966 & 0.971 & 0.968 & 0.973 & 0.961 & 0.966 & 0.941 & 0.975 \\
    (1) vs (3) & 0.798 & 0.825 & 0.817 & 0.748 & 0.811 & 0.813 & 0.763 & 0.828 \\
    (1) vs (4) & 0.959 & 0.96 & 0.768 & 0.805 & 0.967 & 0.948 & 0.737 & 0.973 \\
    (1) vs (5) & 0.939 & 0.946 & 0.774 & 0.817 & 0.942 & 0.933 & 0.745 & 0.955 \\
    (1) vs (6) & 0.791 & 0.819 & 0.774 & 0.728 & 0.8 & 0.804 & 0.691 & 0.823 \\
    (2) vs (3) & 0.78 & 0.815 & 0.8 & 0.741 & 0.793 & 0.8 & 0.754 & 0.817 \\
    (2) vs (4) & 0.935 & 0.944 & 0.754 & 0.785 & 0.939 & 0.934 & 0.727 & 0.956 \\
    (2) vs (5) & 0.963 & 0.965 & 0.776 & 0.814 & 0.971 & 0.956 & 0.757 & 0.975 \\
    (2) vs (6) & 0.774 & 0.81 & 0.766 & 0.721 & 0.78 & 0.792 & 0.692 & 0.812 \\
    (3) vs (4) & 0.793 & 0.828 & 0.681 & 0.689 & 0.805 & 0.817 & 0.67 & 0.827 \\
    (3) vs (5) & 0.773 & 0.815 & 0.672 & 0.677 & 0.787 & 0.8 & 0.666 & 0.813 \\
    (3) vs (6) & 0.995 & 0.999 & 0.872 & 0.954 & 0.975 & 0.997 & 0.846 & 0.998 \\
    (4) vs (5) & 0.966 & 0.975 & 0.958 & 0.962 & 0.959 & 0.973 & 0.957 & 0.976 \\
    (4) vs (6) & 0.791 & 0.825 & 0.745 & 0.728 & 0.803 & 0.813 & 0.74 & 0.825 \\
    (5) vs (6) & 0.771 & 0.812 & 0.721 & 0.704 & 0.782 & 0.796 & 0.712 & 0.811 \\
\end{tabular}
\end{table*}

\begin{figure*}
    \begin{subfigure}{\linewidth}
        \centering
        \includegraphics[width=0.49\linewidth]{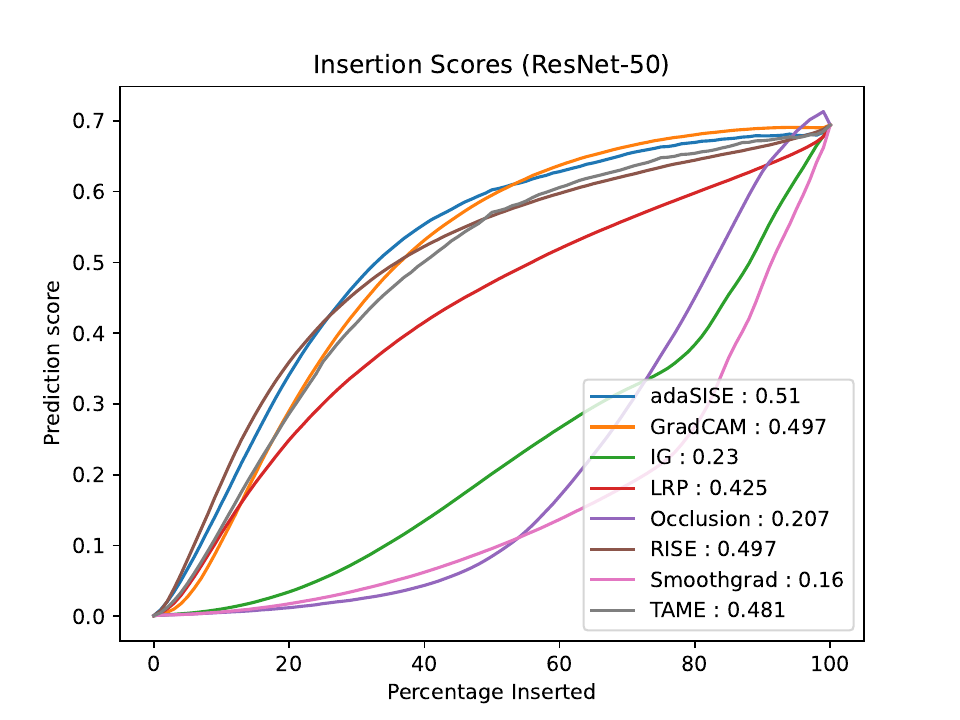}
        \includegraphics[width=0.49\linewidth]{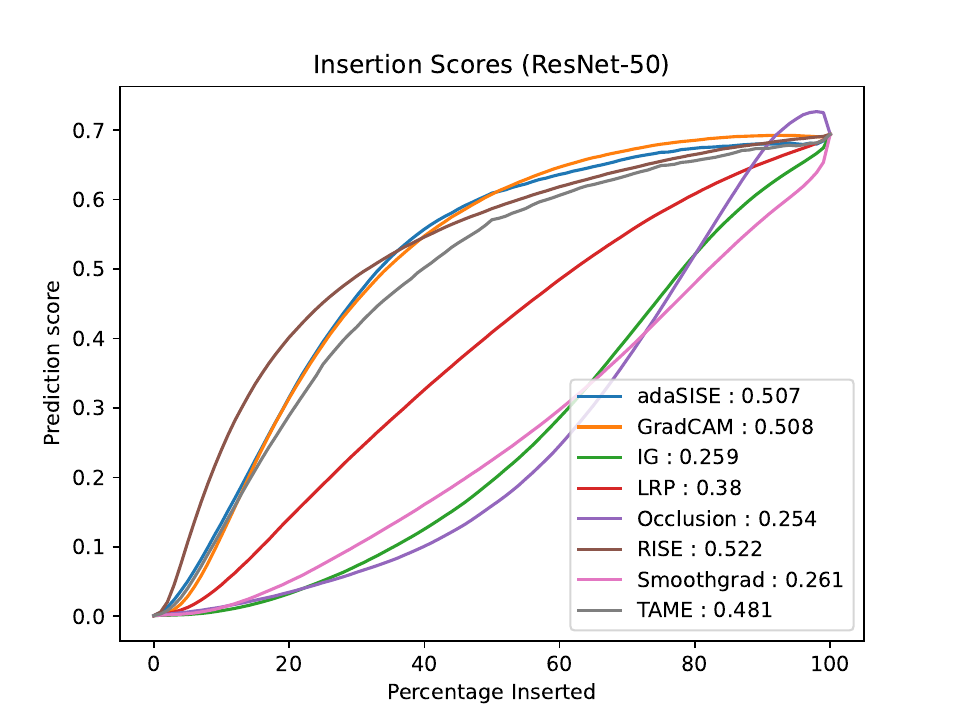}
        \caption{mean-valued baseline}
    \end{subfigure}
    \begin{subfigure}{\linewidth}
        \centering
        \includegraphics[width=0.49\linewidth]{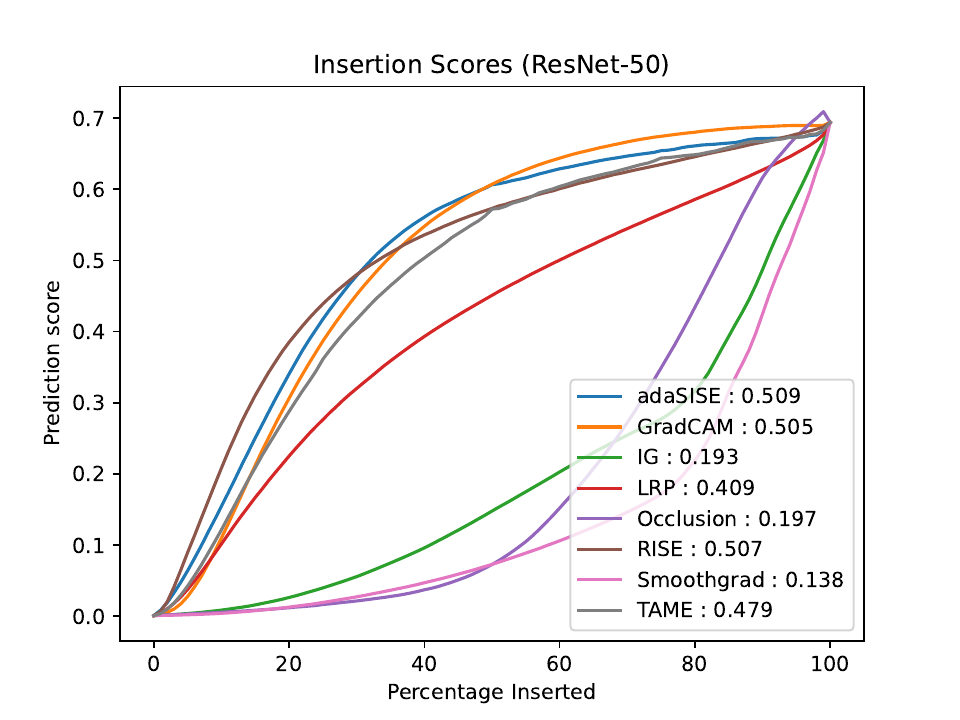}
        \includegraphics[width=0.49\linewidth]{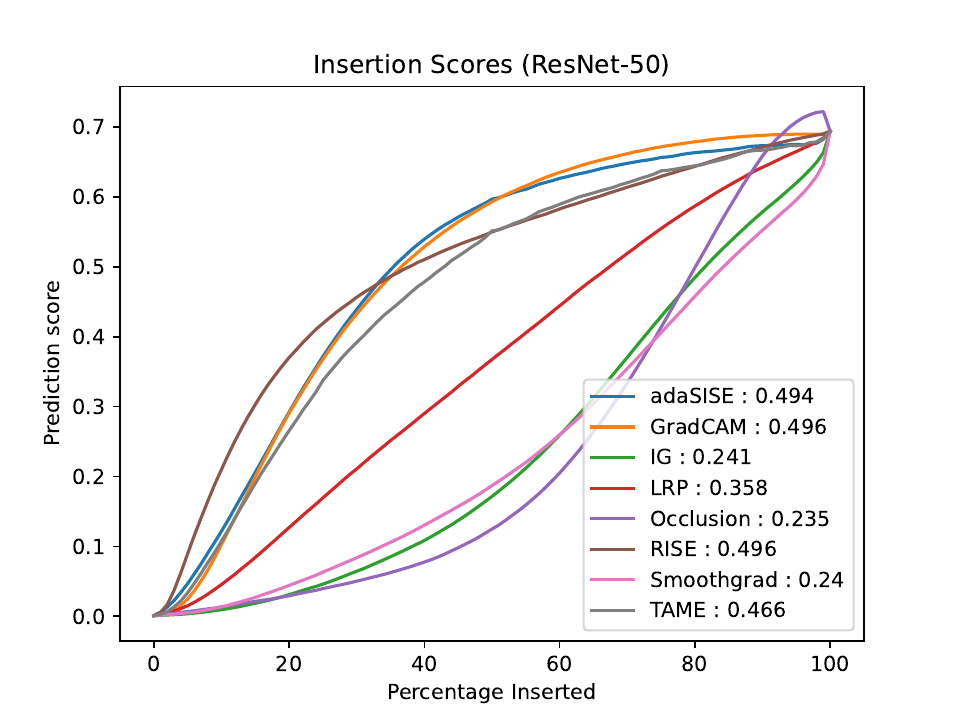}
        \caption{uniform random baseline}
    \end{subfigure}
    \begin{subfigure}{\linewidth}
        \centering
        \includegraphics[width=0.49\linewidth]{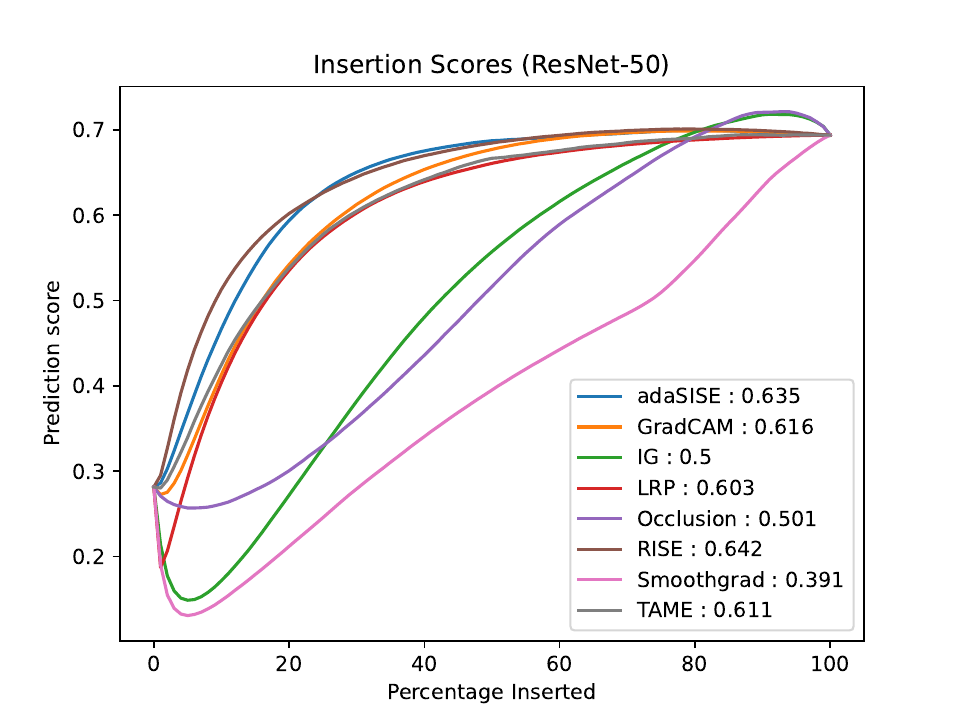}
        \includegraphics[width=0.49\linewidth]{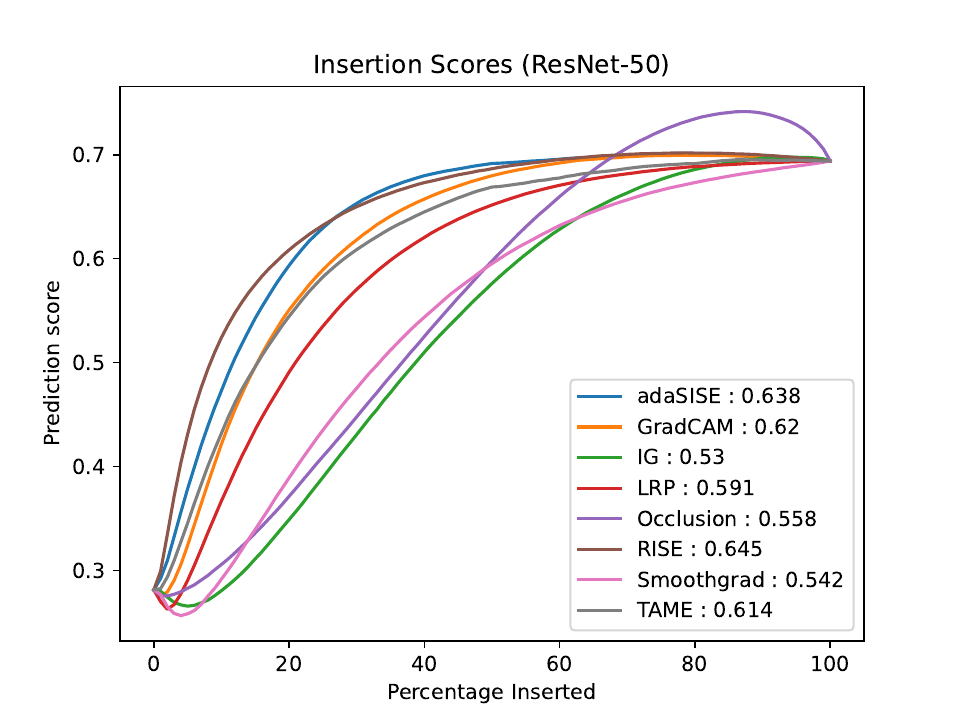}
        \caption{blurred baseline}
    \end{subfigure}
    \caption{The insertion curves for the ResNet-50 network. The curves on the left use pixel-level replacements, while the curves on the right use region-level replacement}
    \label{fig:insert_resnet}
\end{figure*}

\begin{figure*}
    \begin{subfigure}{\linewidth}
        \centering
        \includegraphics[width=0.49\linewidth]{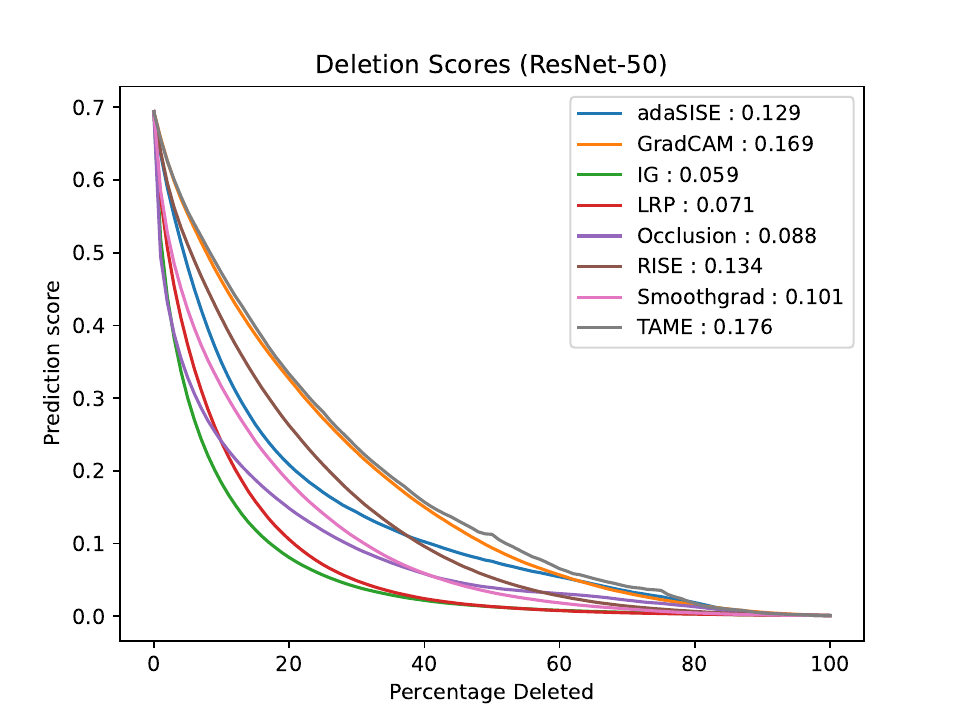}
        \includegraphics[width=0.49\linewidth]{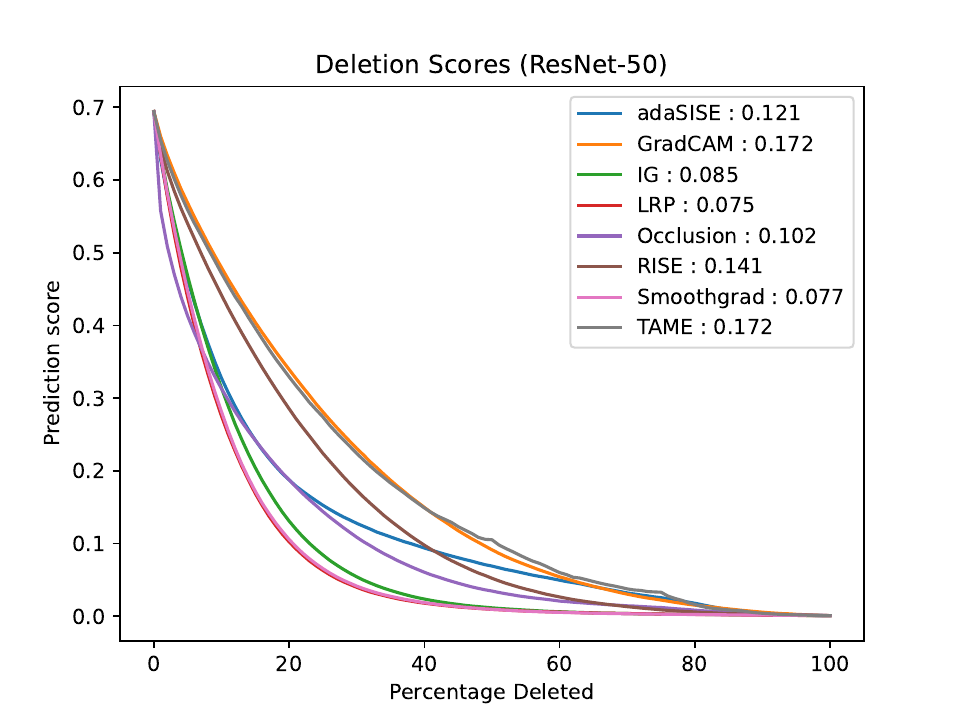}
        \caption{mean-valued baseline}
    \end{subfigure}
    \begin{subfigure}{\linewidth}
        \centering
        \includegraphics[width=0.49\linewidth]{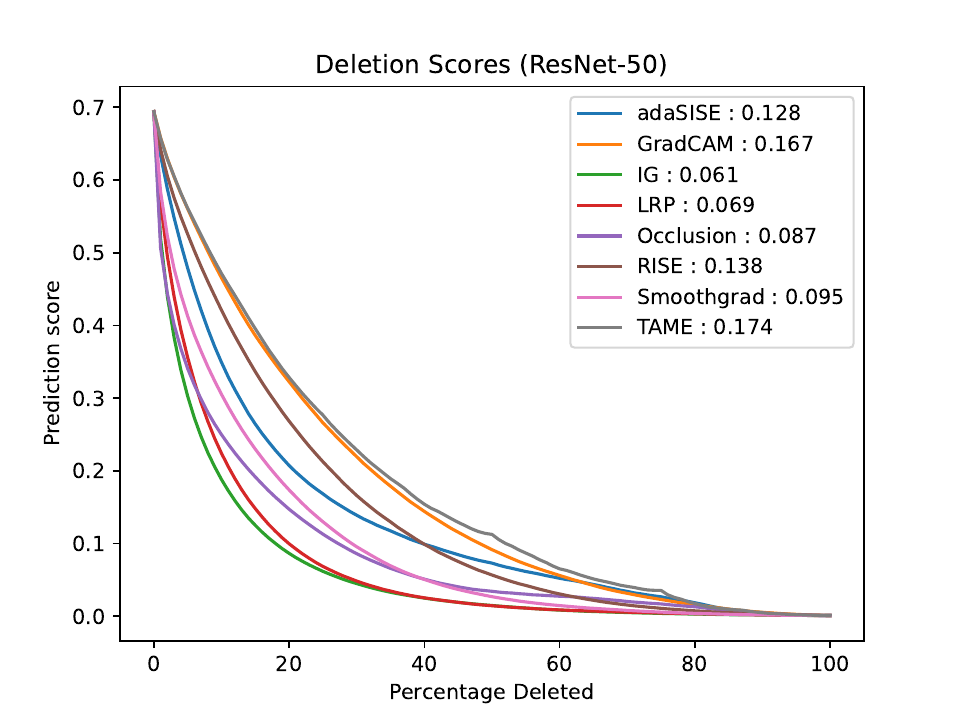}
        \includegraphics[width=0.49\linewidth]{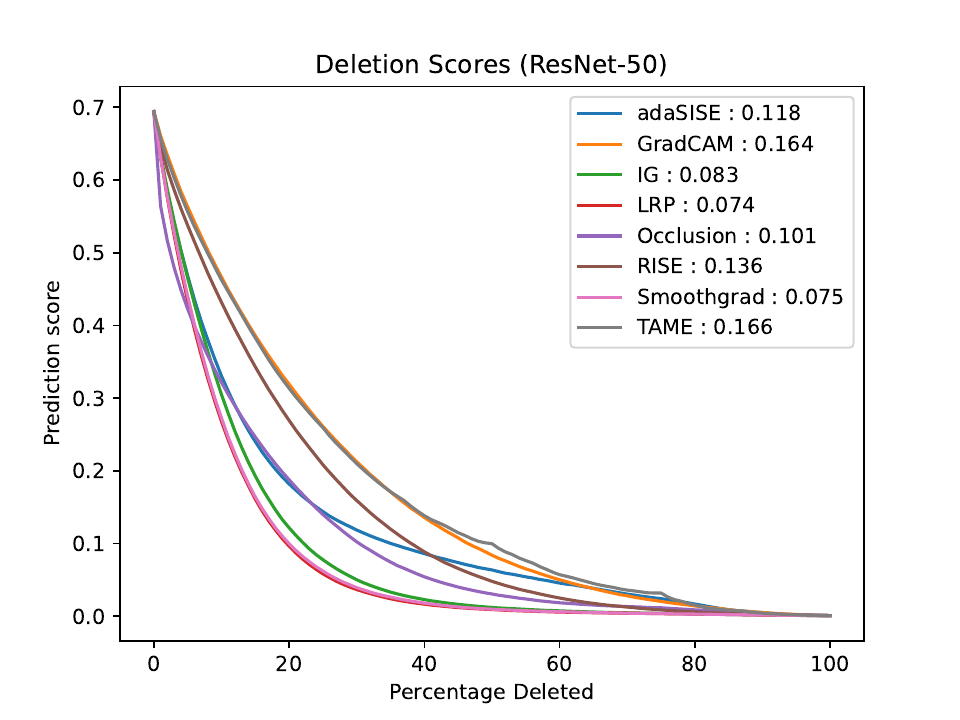}
        \caption{uniform random baseline (over 5 iterations)}
    \end{subfigure}
    \begin{subfigure}{\linewidth}
        \centering
        \includegraphics[width=0.49\linewidth]{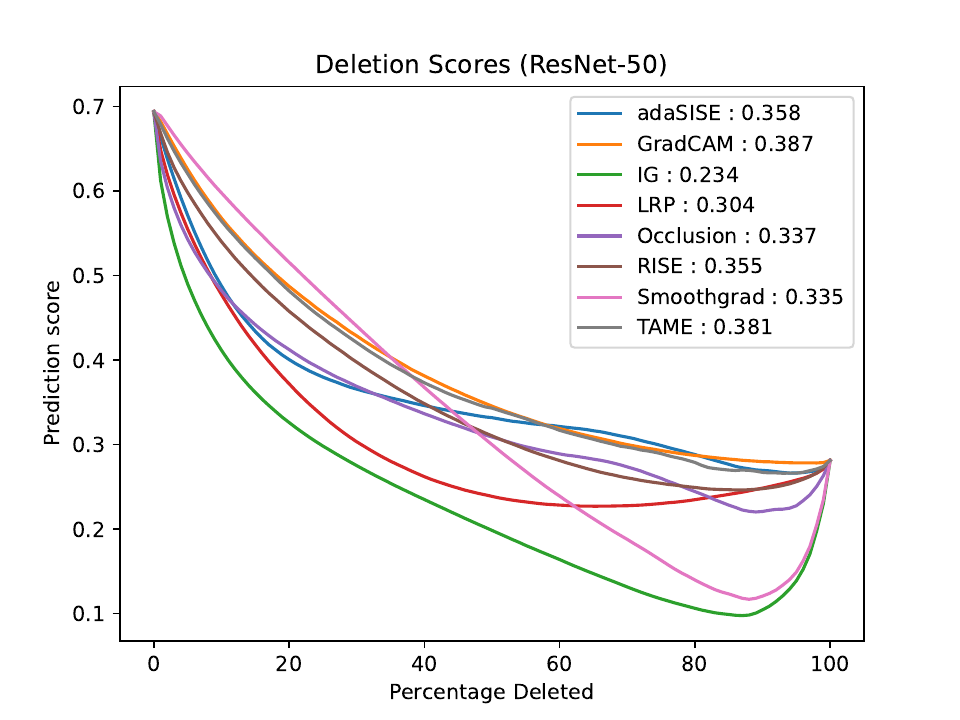}
        \includegraphics[width=0.49\linewidth]{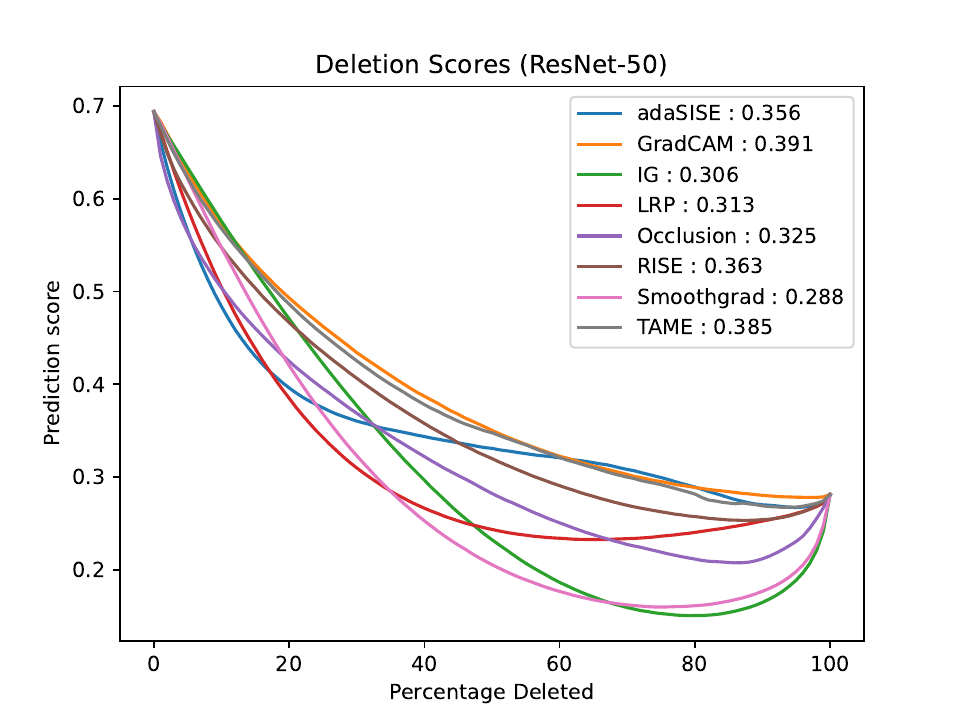}
        \caption{blurred baseline}
    \end{subfigure}
    \caption{The deletion curves for the ResNet-50 network. The curves on the left use pixel-level replacements, while the curves on the right use region-level replacement}
    \label{fig:delete_resnet}
\end{figure*}  

Notice that Integrated Gradients, \textsc{Smoothgrad} and Occlusion have better region-based insertion scores, but slightly worse region-based deletion scores, while the other explanation methods show little to no difference. (See Figures \ref{fig:insert_resnet} -- \ref{fig:delete_resnet}) This is due to the artificial coarsening done by region-based insertion / deletion. By considering regions, when a pixel is disturbed, it is often due to another pixel in the region that has a high relevance. This is similar with coarse explanations where highly relevant pixels are often in close proximity. As such, region-based perturbation follows the reasoning from the main paper where we found that the sparsity of an explanation has an influence on the used evaluation metric. In this case, the explanation is not made coarse by itself, but the evaluation metric achieves similar effect. 


    

We also break the correlation tables that we mentioned in the main paper down in greater detail (\autoref{tab:consistency_resnet}, \autoref{tab:consistency_vgg}), from which we can infer more detailed information. 
Specifically it is once again clear that correlating a blurred baseline with either a mean-valued or uniform random baseline leads to lower correlation scores than correlating a mean-valued baseline with a uniform random baseline. This is due to the fact that blurring removes less information from the input than uniform noise or a mean value does. We can see that the average prediction score of the blurred image is still around 0.3, while for the other methods the prediction score is almost 0. This might impact the ordering in which pixels need to be inserted or deleted for an optimal evaluation score compared to another baseline.
Another observation is that the correlation scores between a pixel-based and a region-based evaluation for the gradient-based explanations are significantly lower than for the other explanations. This indicates that the addition of region-based deletion has a large impact on the evaluation of gradient-based methods, but not on the other methods. Once again, we can see from visual inspection that indeed the region-based insertion and deletion curves for gradient-based methods differ significantly compared to the pixel-based insertion and deletion curves (see Figures \ref{fig:insert_resnet} -- \ref{fig:delete_resnet}).

\section{Uninformative Values}
\label{sec:uninformative}
\begin{figure}
    \begin{subfigure}{0.3\linewidth}
        \includegraphics[width=\linewidth]{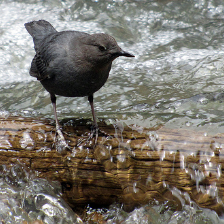}
        \subcaption{}
    \end{subfigure}
    \hspace{2em}
    \begin{subfigure}{0.3\linewidth}
        \includegraphics[width=\linewidth]{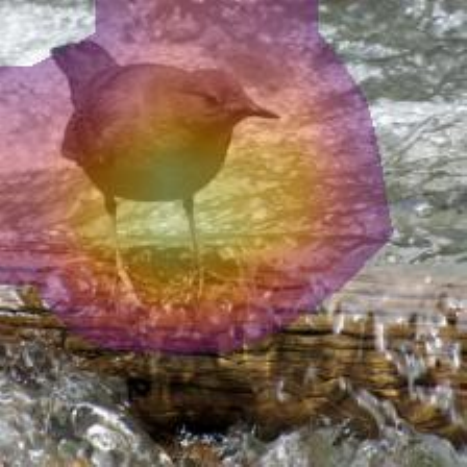}
        \subcaption{}
    \end{subfigure} 
    \hspace{2em}
    \begin{subfigure}{0.3\linewidth}
        \includegraphics[width=\linewidth]{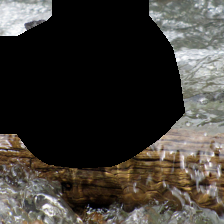}
        \subcaption{}
    \end{subfigure} \\
    \begin{subfigure}{0.3\linewidth}
        \includegraphics[width=\linewidth]{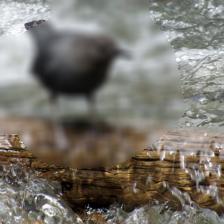}
        \subcaption{}
    \end{subfigure} 
    \hspace{2em}
    \begin{subfigure}{0.3\linewidth}
        \includegraphics[width=\linewidth]{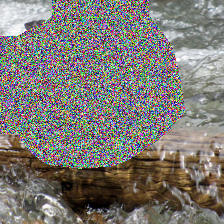}
        \subcaption{}
    \end{subfigure}
    \hspace{2em}
    \begin{subfigure}{0.3\linewidth}
        \includegraphics[width=\linewidth]{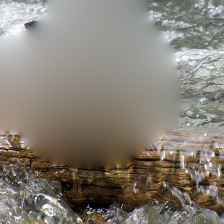}
        \subcaption{}
    \end{subfigure}
    \caption{Different types of uninformative values used within the Deletion procedure. From left-to-right, top-to-bottom: (a) The original image, (b) Grad-CAM based explanation, (c) mean baseline, (d) gaussian blur, (e) uniform noise, (f) imputation via sparse linear functions.}
    \label{fig:uninformative}
\end{figure}

As noted in the main paper, the model-based evaluation methods function by replacing parts of the image with uninformative values. In the literature, there are a number of uninformative values used, namely using black pixels (zero-valued), uniform random noise, blurred pixels and sparse linear equations (ROAD). Each of these follow different intuitions. Zero-valued replacement, uses the dataset mean as that should occur most. Uniform randomness is used to minimize the occurrence of hard edges within the perturbed image. Blurring is often use during the insertion procedure for exactly the same reason, avoiding hard edges. Finally, Sparse Linear Equations are introduced by ROAD to avoid information leakage from the explanation map and are based on the intuition that neighbouring pixels are often highly correlated. A visualisation of the different types is given in \autoref{fig:uninformative} using the deletion procedure where 50\% of the image is perturbed based on the most relevant pixels according to the Grad-CAM explanation method.,

\section{Sparsity in Explanations}
\label{sec:sparseness}
\begin{figure}[H]
    \centering
    \begin{subfigure}{0.24\linewidth}
    \includegraphics[width=0.8\linewidth]{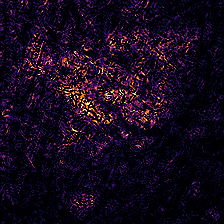}
    \includegraphics[width=0.8\linewidth]{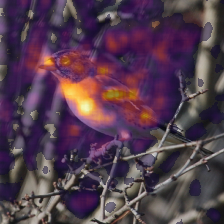}
    \subcaption{Integrated Gradients} \label{fig:coarse_gradients_a}
    \end{subfigure}
    \begin{subfigure}{0.24\linewidth}
    \includegraphics[width=0.8\linewidth]{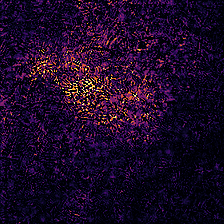}
    \includegraphics[width=0.8\linewidth]{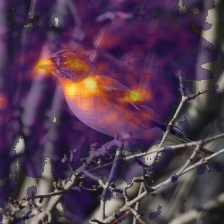}
    \subcaption{\textsc{Smoothgrad}} \label{fig:coarse_gradients_b}
    \end{subfigure}
    \caption{The effect of blurring on the explanations generated by Integrated Gradients and \textsc{Smoothgrad}}
\end{figure}

Here we visualise the result of the blurring operation on the Integrated Gradients (\autoref{fig:coarse_gradients_a}) and \textsc{Smoothgrad} (\autoref{fig:coarse_gradients_b}) explanation methods. 

\section{Binarized Average Drop \%}
\label{sec:drop}

\begin{table}[H]
    \centering
    \caption{The evaluation scores for Average Drop \% with different binarizations.}
    \begin{tabular}{c|c|c|c}
     & orig & 50\% & 75\% \\ \hline \hline
    adaSISE & 47.05\% & 24.77\% & 41.49\% \\
    Grad-CAM & 14.96\% & 22.20\% & 51.11\% \\
    IG & 96.85\% & 71.00\% & 91.21\% \\
    LRP & 66.26\% & 39.8\% & 62.46\% \\
    Occlusion & 94.77\% & 87.46\% & 95.90\% \\
    RISE & 14.02\% & 25.82\% & 48.23\% \\
    \textsc{smoothgrad} & 96.99\% & 84.43\% & 94.63\% \\
    TAME & 27.45\% & 28.33\% & 49.41\% \\
    \end{tabular}
    \label{tab:avg_drop_binary}
\end{table}

In this section, we further study the difference between average drop with real-valued explanations and average drop with binary-valued explanations. We posit that using binary explanation are better for evaluation, as multiplying with a real-valued explanation can introduce unexpected effects. Examples are f.e. colour difference, which can lead to misattribution when combining the explanation and input to calculate the average drop \% metric. 
We show the difference in evaluation score between using real-valued saliency maps and binary saliency maps at different thresholds in \autoref{tab:avg_drop_binary}. Binarizing a saliency map is done on a per-image base by thresholding using a percentile-based approach. Other approaches, such as thresholding on a fixed value are also possible, but not explored during this small experiment. We should additionally note that the inter-method reliability between each level of binarization and real-valued explanations is exactly 1, which means that binarizing doesn't influence the ordering of the scoring.

When correlating the evaluation scores for binarised explanations to the other metrics (\autoref{tab:avg_drop_correl}), we can notice that the phenomenon described in the main paper --- where the correlation coefficients differ wildly between IG and \textsc{Smoothgrad}, and the other explanation methods --- is not present for the lower binarization thresholds. Furthermore, the correlation scores are generally more coherent when using binarization, even at unrealistic thresholds. This further suggest that due to the definition of average drop \% in the literature, some explanation methods --- more specifically those with a few extreme outliers such as Integrated Gradients and \textsc{Smoothgrad} --- are penalized too harsh. Finally, the fact that the correlation scores with the pointing game metric are still very low for different levels of binarization, further reinforces our reasoning that pointing game is no good evaluation metric.

\begin{table*}
    \caption{The correlation scores between the (binarized) Average drop \% metric at different levels of binarization and other metrics}
    \begin{subtable}{\linewidth}
    \centering
    \begin{tabular}{c|c|c|c|c|c|c|c}
    & real-valued & 50\% & 70\% & 75\% & 80\% & 85\% & 90\% \\ \hline \hline
    adaSISE & 0.0898 & 0.1361 & 0.1452 & 0.1457 & 0.1461 & 0.131 & 0.0688 \\
    Grad-CAM & 0.1725 & 0.2094 & 0.2182 & 0.2075 & 0.1771 & 0.1159 & 0.0169 \\
    IG & -0.3492 & 0.2411 & 0.1194 & 0.0725 & 0.02 & -0.0416 & -0.1097 \\
    LRP & 0.101 & 0.2398 & 0.1957 & 0.1745 & 0.1457 & 0.106 & 0.0414 \\
    Occlusion & -0.322 & 0.0169 & -0.1445 & -0.1869 & -0.2289 & -0.2655 & -0.3002 \\
    RISE & 0.0167 & 0.1328 & 0.1388 & 0.1355 & 0.126 & 0.1057 & 0.0607 \\
    \textsc{smoothgrad} & -0.3504 & 0.2285 & 0.0934 & 0.0491 & -0.0018 & -0.0607 & -0.128 \\
    TAME & 0.2433 & 0.2645 & 0.2806 & 0.2744 & 0.2572 & 0.218 & 0.1334 \\
    \end{tabular}
    \subcaption{Correlation scores between the Insertion metric and the average drop \% metric.}
    \end{subtable}
    \begin{subtable}{\linewidth}
    \centering

    \begin{tabular}{c|c|c|c|c|c|c|c}
    & real-valued & 50\% & 70\% & 75\% & 80\% & 85\% & 90\% \\ \hline \hline
    adaSISE & 0.0078 & -0.0344 & -0.057 & -0.0503 & -0.0376 & 0.0025 & 0.0854 \\
    Grad-CAM & -0.0626 & -0.1615 & -0.124 & -0.1012 & -0.0655 & -0.0107 & 0.0697 \\
    IG & 0.3182 & -0.2626 & -0.1144 & -0.0632 & -0.0065 & 0.0594 & 0.1332 \\
    LRP & -0.1455 & -0.2617 & -0.2047 & -0.176 & -0.1412 & -0.094 & -0.02 \\
    Occlusion & 0.3091 & 0.0916 & 0.1985 & 0.2284 & 0.2559 & 0.2755 & 0.2965 \\
    RISE & 0.1059 & -0.0782 & -0.0826 & -0.0807 & -0.0767 & -0.0571 & -0.0127 \\
    \textsc{smoothgrad} & 0.2725 & -0.3762 & -0.174 & -0.1077 & -0.03 & 0.058 & 0.1546 \\
    TAME & 0.0419 & -0.0282 & -0.0334 & -0.0257 & -0.0101 & 0.0199 & 0.0787 \\
    
    \end{tabular}
    \subcaption{Correlation scores between the Deletion metric and the average drop \% metric.}
    \end{subtable}
    \begin{subtable}{\linewidth}
    \centering
    \begin{tabular}{c|c|c|c|c|c|c|c}
  & real-valued & 50\% & 70\% & 75\% & 80\% & 85\% & 90\% \\ \hline \hline
adaSISE & -0.0251 & -0.0422 & -0.0387 & -0.035 & -0.0342 & -0.0291 & -0.01 \\
Grad-CAM & -0.032 & -0.0309 & -0.0144 & -0.0067 & 0.0026 & 0.0192 & 0.0366 \\
IG & 0.0673 & -0.0278 & -0.0144 & -0.0074 & 0.0023 & 0.0128 & 0.0231 \\
LRP & -0.0487 & -0.1528 & -0.1412 & -0.1319 & -0.1184 & -0.1048 & -0.0778 \\
Occlusion & 0.0606 & -0.0013 & 0.0344 & 0.0441 & 0.0521 & 0.0595 & 0.0652 \\
RISE & -0.091 & -0.0838 & -0.0688 & -0.0659 & -0.0565 & -0.0414 & -0.0176 \\
\textsc{smoothgrad} & 0.0665 & -0.0227 & -0.0052 & 0.004 & 0.0104 & 0.019 & 0.0272 \\
TAME & -0.2752 & -0.2763 & -0.2058 & -0.1853 & -0.1598 & -0.1262 & -0.0784 \\
    \end{tabular}
    \subcaption{Correlation scores between the Pointing game metric and the average drop \% metric.}
    \end{subtable}
    \label{tab:avg_drop_correl}
\end{table*}

\section{VGG16 Results}
\label{sec:vgg16}

\begin{figure*}
    \centering
    \begin{subfigure}{\linewidth}
        \centering
        \includegraphics[width=0.49\linewidth]{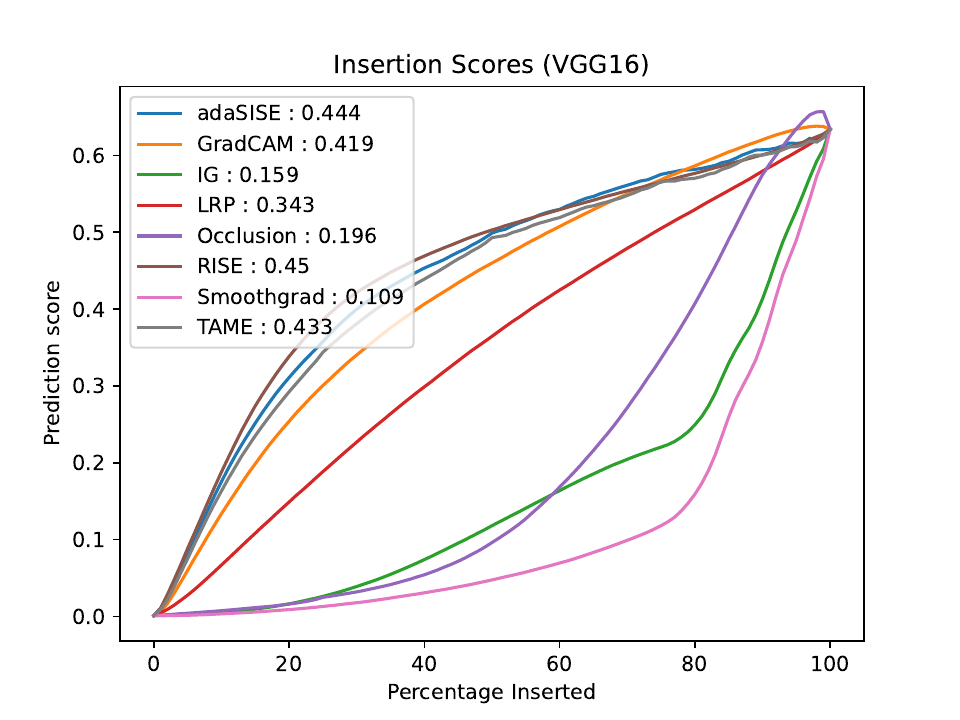}
        \includegraphics[width=0.49\linewidth]{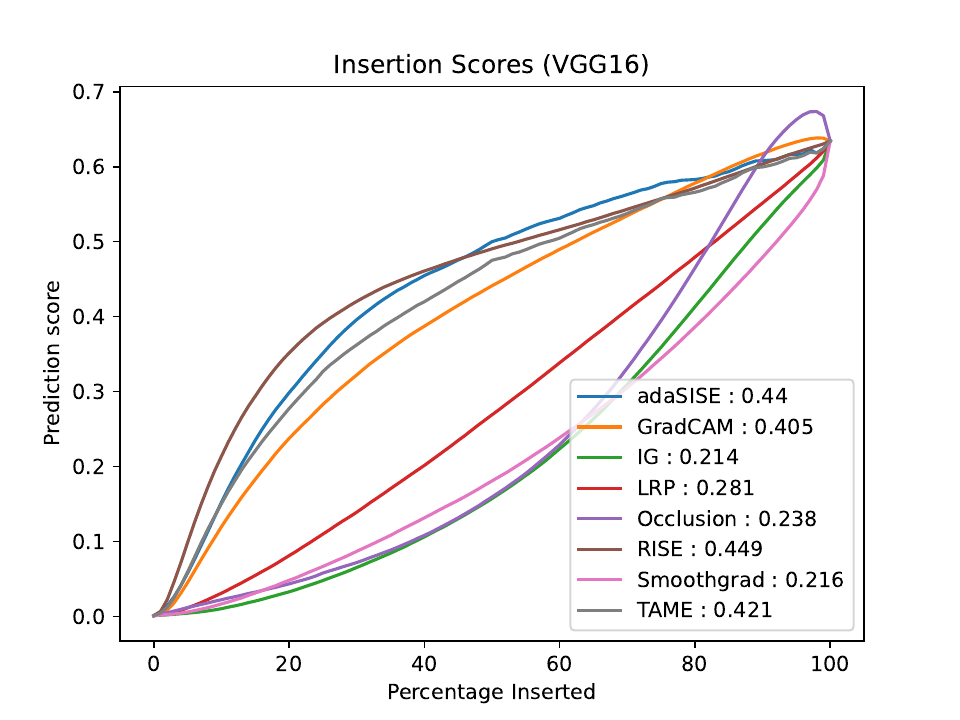}
        \caption{mean-valued baseline}
    \end{subfigure}
    \begin{subfigure}{\linewidth}
        \centering
        \includegraphics[width=0.49\linewidth]{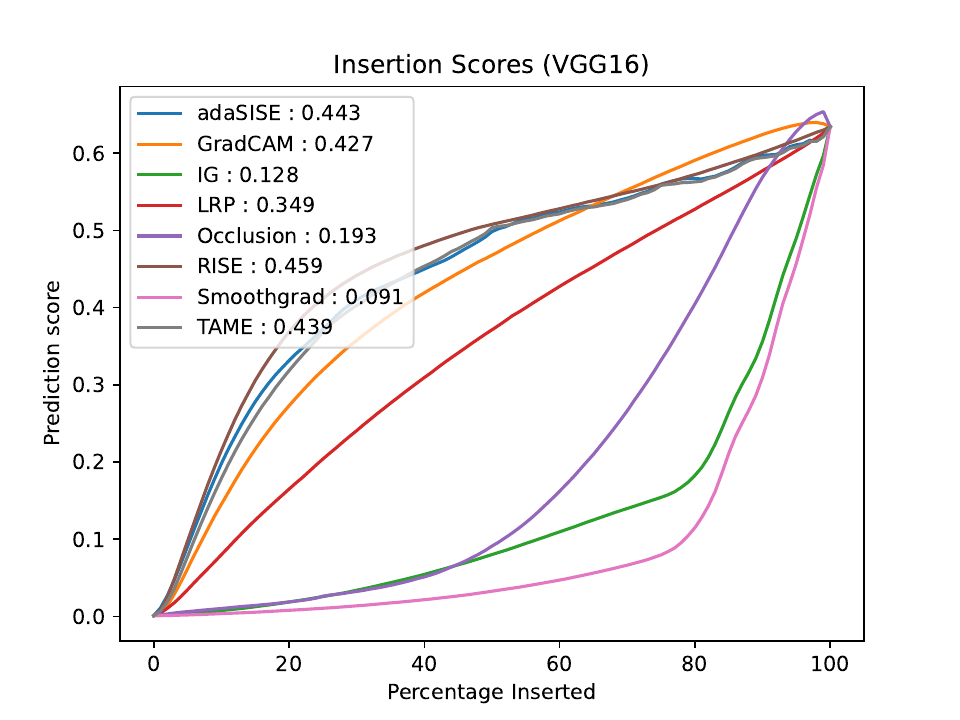}
        \includegraphics[width=0.49\linewidth]{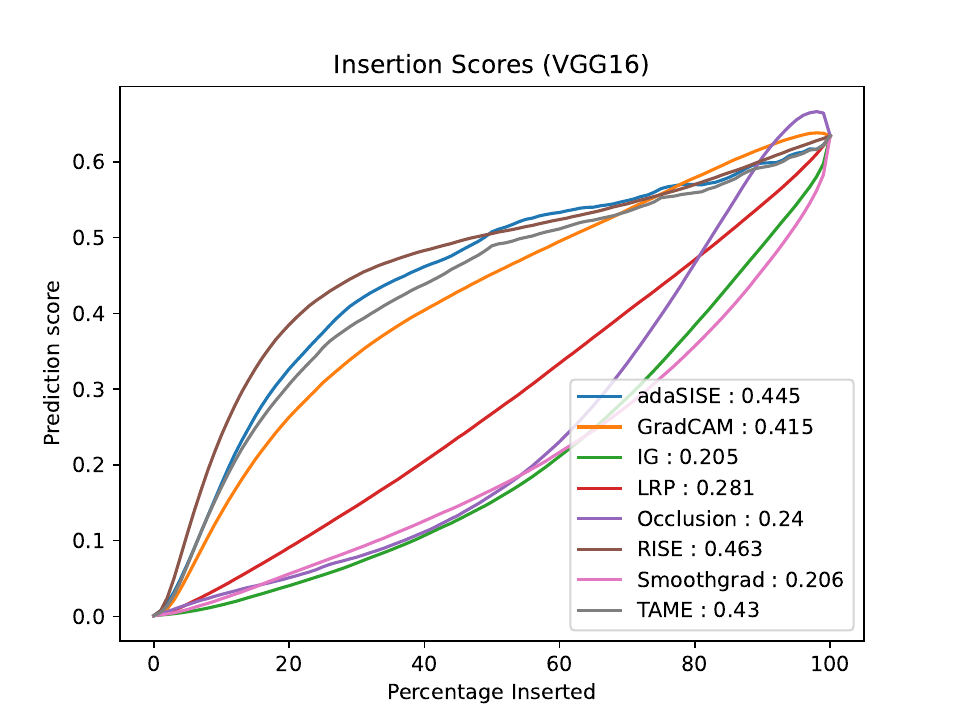}
        \caption{uniform random baseline (over 5 iterations)}
    \end{subfigure}
    \begin{subfigure}{\linewidth}
        \centering
        \includegraphics[width=0.49\linewidth]{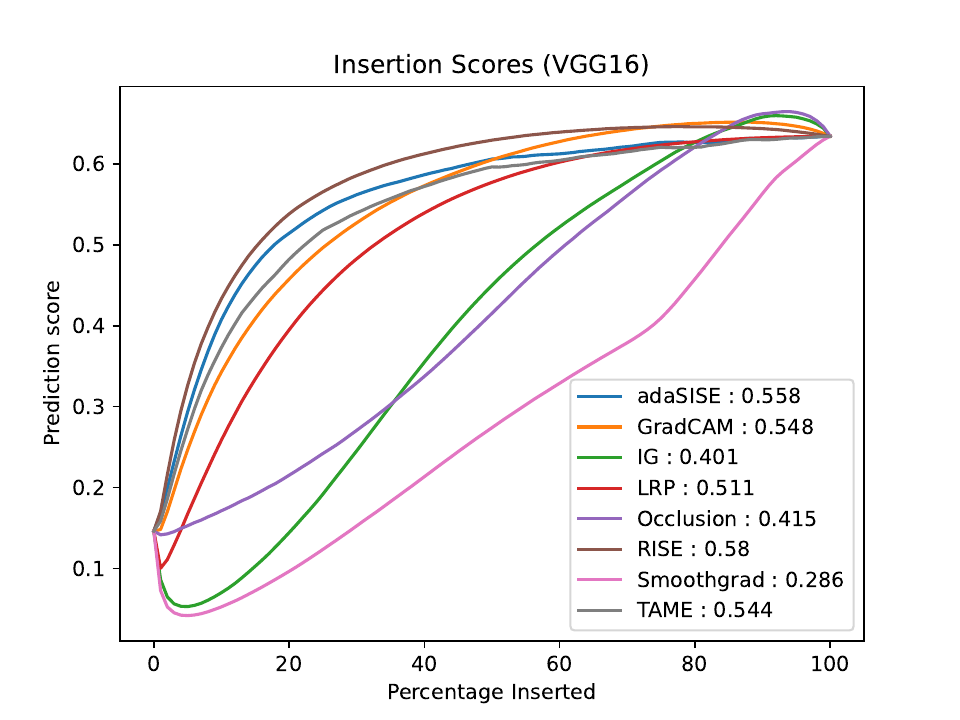}
        \includegraphics[width=0.49\linewidth]{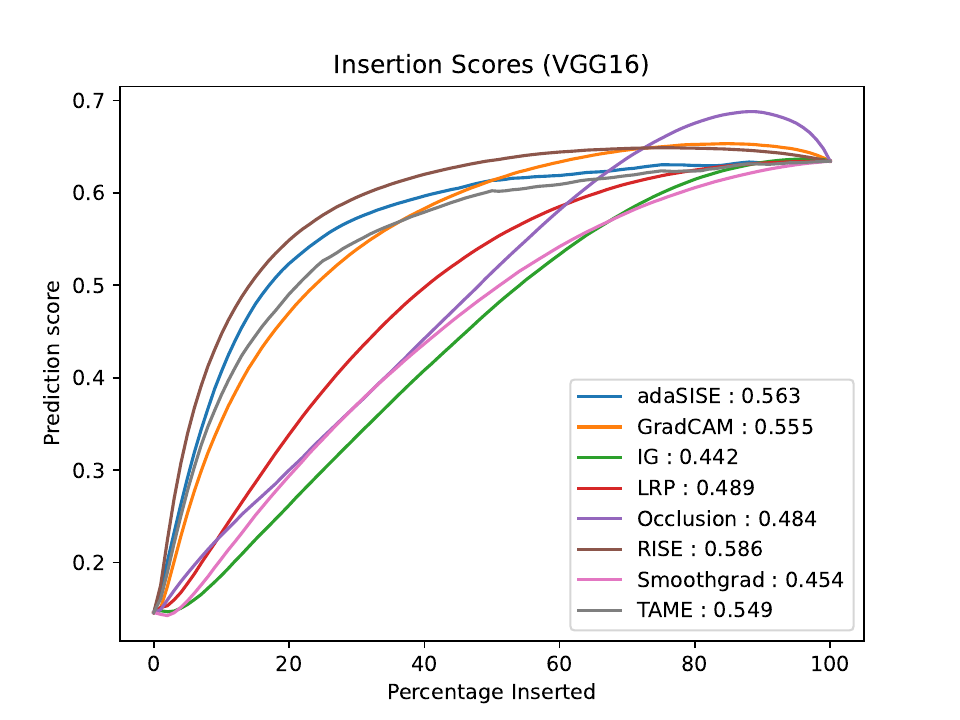}
        \caption{blurred baseline}
    \end{subfigure}
    \caption{The insertion curves for the VGG16 network. The curves on the left use pixel-level replacements, while the curves on the right use region-level replacement}
    \label{fig:insert_vgg}
\end{figure*}

\begin{figure*}
    \begin{subfigure}{\linewidth}
        \centering
        \includegraphics[width=0.49\linewidth]{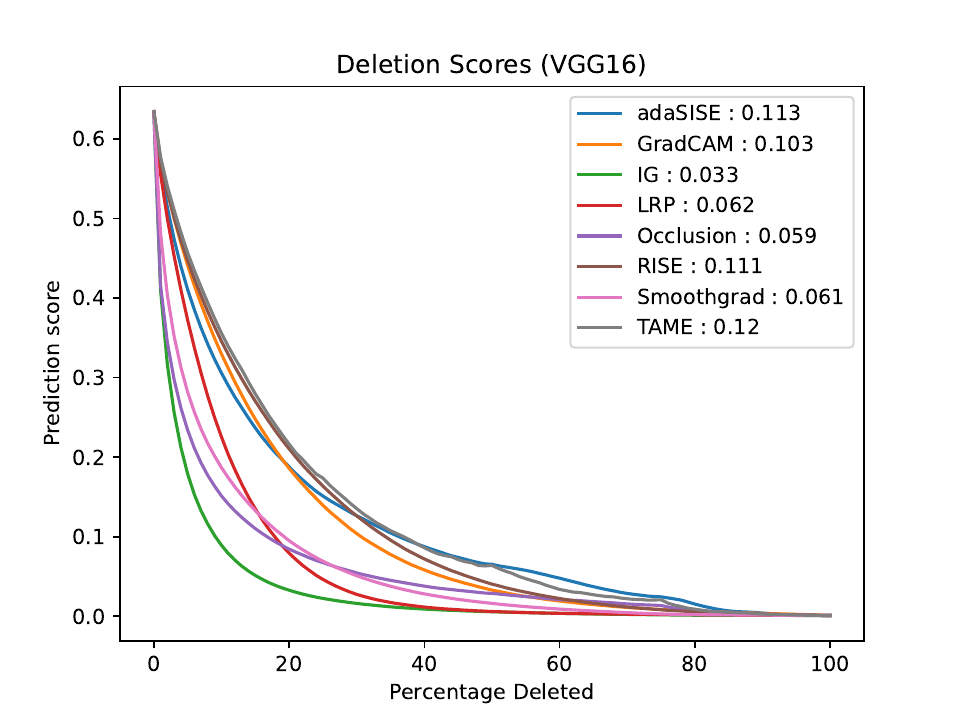}
        \includegraphics[width=0.49\linewidth]{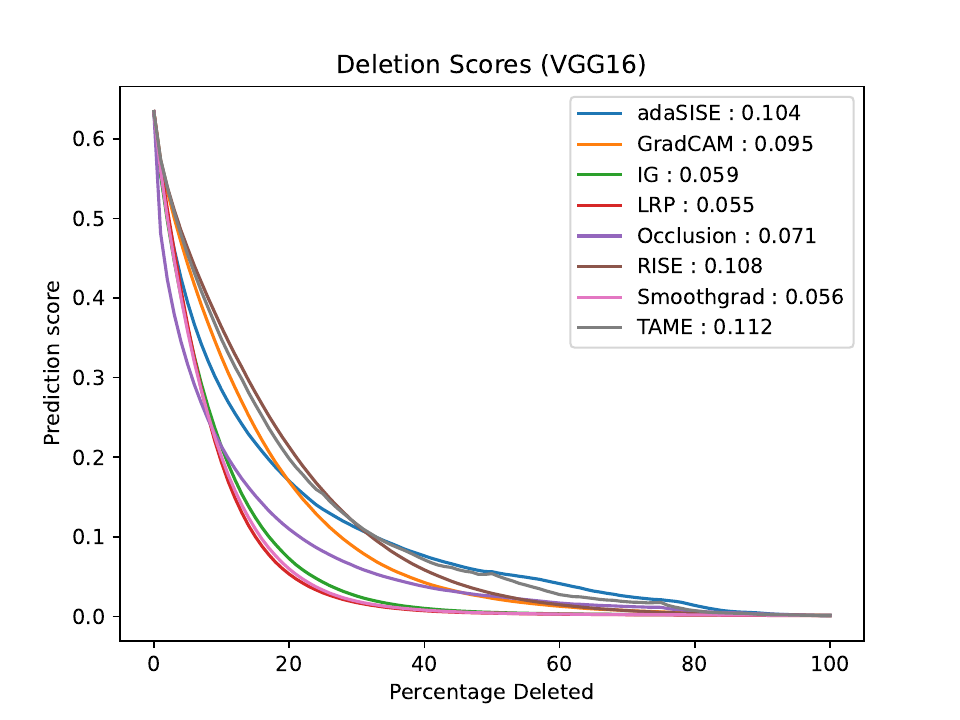}
        \caption{mean-valued baseline}
    \end{subfigure}
    \begin{subfigure}{\linewidth}
        \centering
        \includegraphics[width=0.49\linewidth]{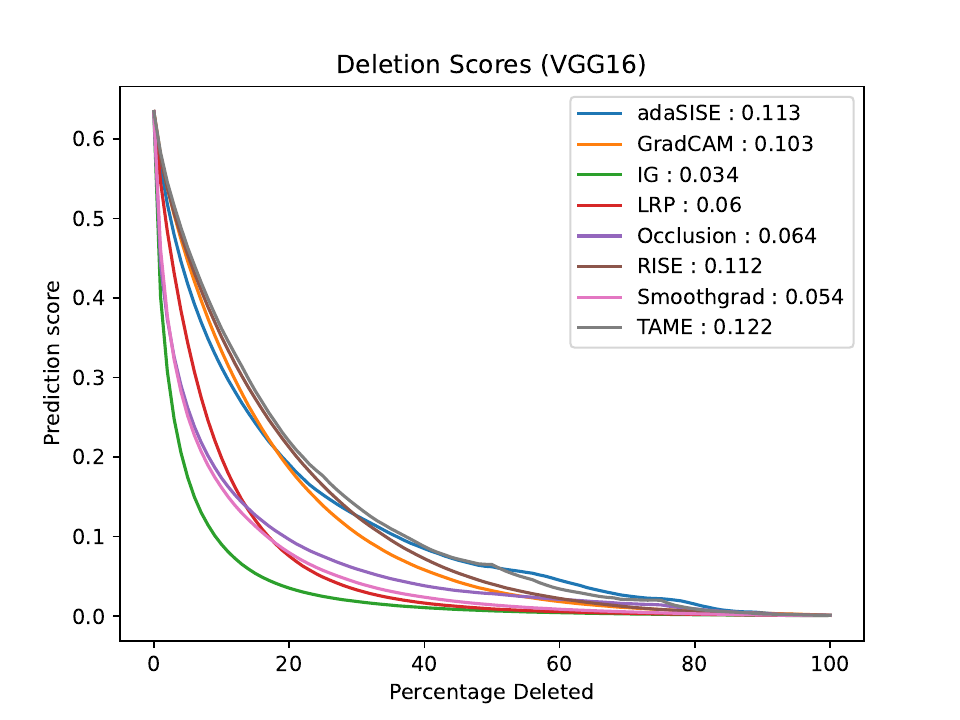}
        \includegraphics[width=0.49\linewidth]{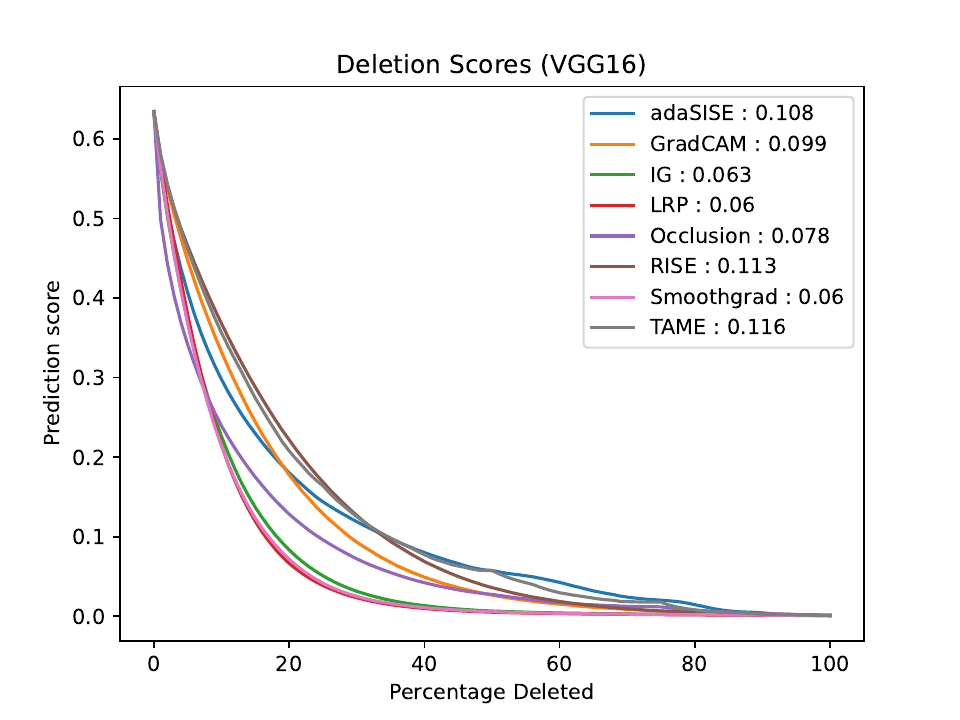}
        \caption{uniform random baseline (over 5 iterations)}
    \end{subfigure}
    \begin{subfigure}{\linewidth}
        \centering
        \includegraphics[width=0.49\linewidth]{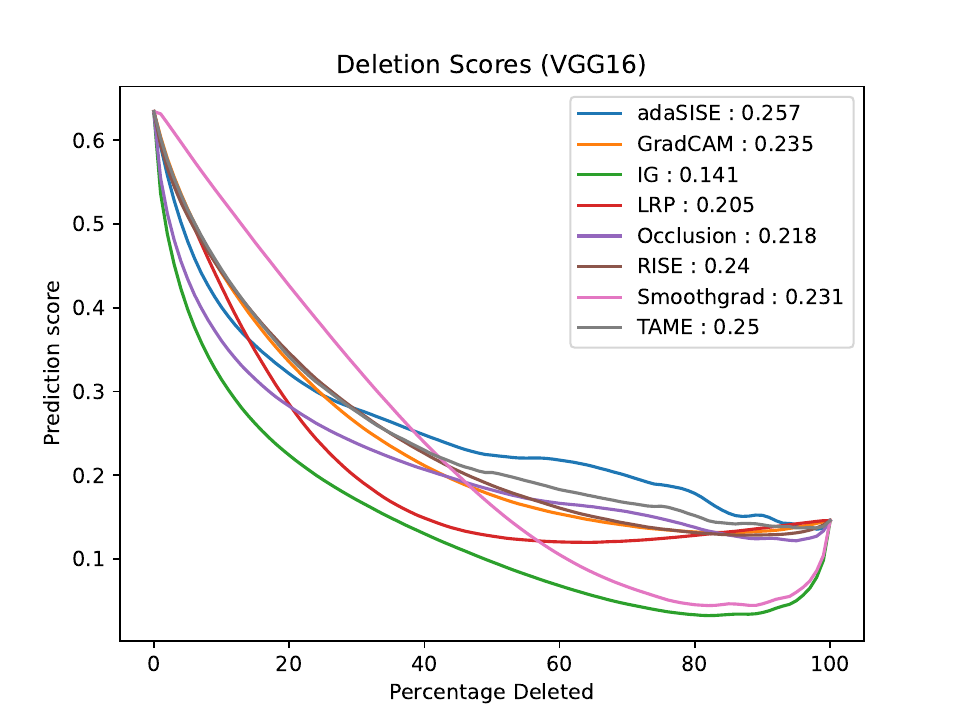}
        \includegraphics[width=0.49\linewidth]{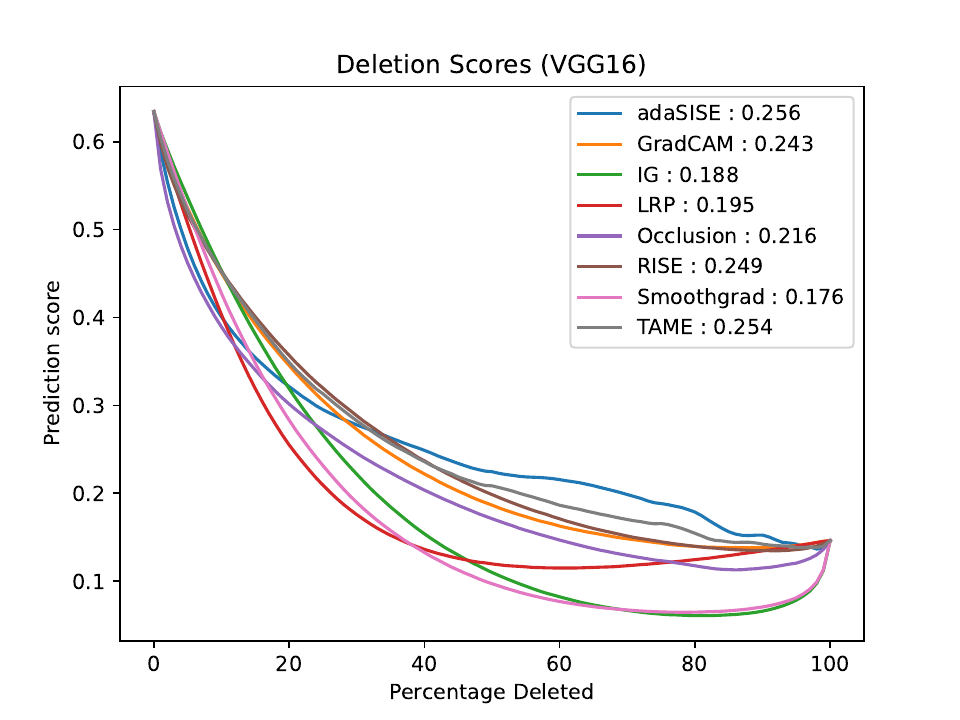}
        \caption{blurred baseline}
    \end{subfigure}
    \caption{The deletion curves for the VGG16 network. The curves on the left use pixel-level replacements, while the curves on the right use region-level replacement}
    \label{fig:delete_vgg}
\end{figure*}

\begin{figure*}
\centering
\includegraphics[width=0.45\linewidth]{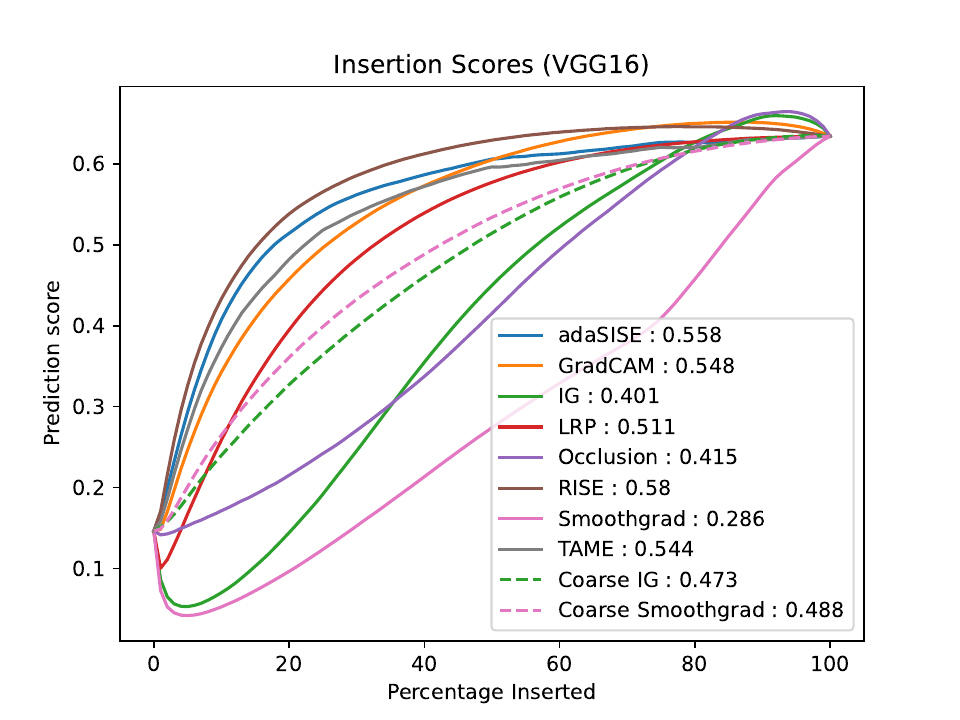}
\includegraphics[width=0.45\linewidth]{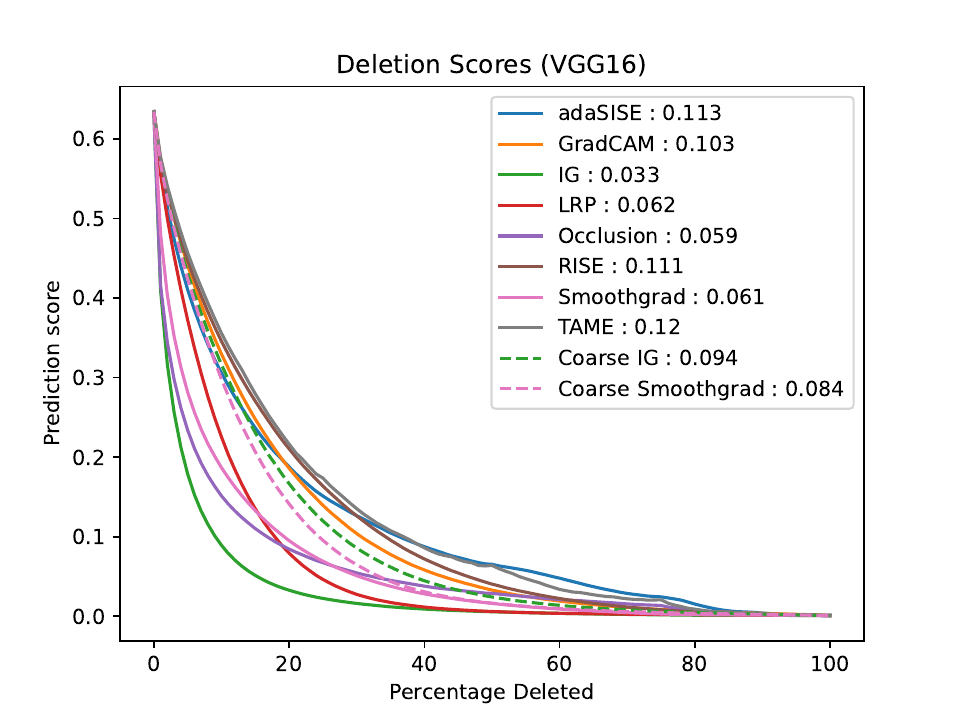}

\begin{tabular}{c|c|c|c|c|c|c|c|c}
& adaSISE & Grad-CAM  & IG (\textit{*})& LRP & Occlusion & RISE & \textsc{smoothgrad} (\textit{*}) & TAME \\ \hline \hline 
Pointing ($\uparrow$) & 90.11\% & 89.21\% & 86.57\% {(\footnotesize86.85\%)}& 84.80\% & 84.52\% & \textbf{91.41\%} & 87.92\% {\footnotesize(88.66\%)} & 87.25\% \\ \hline
Drop \% ($\downarrow$) & 59.50\% & 26.44\% & 51.70\% {\footnotesize(96.46\%)} & 85.14\% & 94.05\% & \textbf{13.53\%} & 51.05\% {\footnotesize(96.52\%)} & 42.37\%\\ \hline
I.i.C. ($\uparrow$) & 14.18\% & 29.87\% & 17.66\% {\footnotesize(1.93\%)} & 4.97\% & 2.93 \% & \textbf{44.55\%} & 14.72\% {\footnotesize(1.91\%)} & 21.90\% \\ \hline
ROAD ($\downarrow$) & 0.148 & 0.136 & 0.155  {\footnotesize(0.202)} & \textbf{0.135} & 0.166 & 0.367 & 0.141 {\footnotesize(0.136)} & 0.148\\ \hline
\end{tabular}

\caption{Comparing blurred versions of IG and \textsc{smoothgrad} explanations to the other explanation methods. Top: The insertion and deletion curves. Bottom : the other evaluation results in tabular form (Original values of IG and \textsc{smoothgrad} are provided between brackets).}
\label{fig:insert_coarse_grads_vgg}
\end{figure*}

Below, we list the visualizations and tables associated with VGG16. Overall, we find the same conclusions as on ResNet-50. The following tables and figures are provided: Detailed internal consistency results (\autoref{tab:consistency_vgg}), a figure describing the effect of blurring on IG and \textsc{smoothgrad} (\autoref{fig:insert_coarse_grads_vgg}), and insertion + deletion curves for the different configurations (\autoref{fig:insert_vgg} and \autoref{fig:delete_vgg}).

\begin{table*}
    \centering
    \caption{The correlations between different configurations of the insertion / deletion metric for the VGG16 network, broken down by explanation method. (1) = mean+pixel, (2) = uniform+pixel, (3) = blur+pixel, (4) = mean+region, (5) = uniform+region, (6) = blur+region}
    \label{tab:consistency_vgg}
    \begin{tabular}{c|c|c|c|c|c|c|c|c}
    \textbf{insertion} & adaSISE & Grad-CAM & IG & LRP & Occlusion & RISE & \textsc{smoothgrad} & TAME\\ \hline \hline
    (1) vs (2) & 0.966 & 0.964 & 0.941 & 0.97 & 0.936 & 0.962 & 0.959 & 0.972 \\
    (1) vs (3) & 0.882 & 0.875 & 0.747 & 0.885 & 0.779 & 0.884 & 0.808 & 0.892 \\
    (1) vs (4) & 0.964 & 0.95 & 0.689 & 0.92 & 0.917 & 0.939 & 0.711 & 0.97 \\
    (1) vs (5) & 0.943 & 0.934 & 0.707 & 0.918 & 0.879 & 0.926 & 0.724 & 0.954 \\
    (1) vs (6) & 0.873 & 0.864 & 0.634 & 0.868 & 0.766 & 0.872 & 0.714 & 0.886 \\
    (2) vs (3) & 0.861 & 0.853 & 0.74 & 0.874 & 0.745 & 0.861 & 0.81 & 0.879 \\
    (2) vs (4) & 0.945 & 0.937 & 0.676 & 0.901 & 0.887 & 0.925 & 0.691 & 0.954 \\
    (2) vs (5) & 0.974 & 0.964 & 0.716 & 0.931 & 0.93 & 0.956 & 0.718 & 0.978 \\
    (2) vs (6) & 0.855 & 0.847 & 0.65 & 0.852 & 0.74 & 0.853 & 0.718 & 0.874 \\
    (3) vs (4) & 0.87 & 0.845 & 0.703 & 0.841 & 0.748 & 0.853 & 0.697 & 0.874 \\
    (3) vs (5) & 0.856 & 0.834 & 0.699 & 0.832 & 0.724 & 0.844 & 0.695 & 0.865 \\
    (3) vs (6) & 0.994 & 0.996 & 0.808 & 0.976 & 0.959 & 0.996 & 0.832 & 0.996 \\
    (4) vs (5) & 0.969 & 0.97 & 0.955 & 0.962 & 0.942 & 0.969 & 0.959 & 0.975 \\
    (4) vs (6) & 0.87 & 0.845 & 0.815 & 0.847 & 0.772 & 0.853 & 0.803 & 0.873 \\
    (5) vs (6) & 0.856 & 0.834 & 0.78 & 0.829 & 0.748 & 0.844 & 0.771 & 0.864 \\

    \end{tabular}
    \begin{tabular}{c|c|c|c|c|c|c|c|c}
     \textbf{deletion} & adaSISE & Grad-CAM & IG & LRP & Occlusion & RISE & \textsc{smoothgrad} & TAME\\ \hline \hline
    (1) vs (2) & 0.959 & 0.957 & 0.97 & 0.964 & 0.961 & 0.959 & 0.947 & 0.964 \\
    (1) vs (3) & 0.847 & 0.852 & 0.774 & 0.727 & 0.824 & 0.864 & 0.747 & 0.859 \\
    (1) vs (4) & 0.956 & 0.949 & 0.757 & 0.849 & 0.969 & 0.944 & 0.721 & 0.966 \\
    (1) vs (5) & 0.939 & 0.935 & 0.772 & 0.861 & 0.948 & 0.937 & 0.737 & 0.949 \\
    (1) vs (6) & 0.838 & 0.84 & 0.761 & 0.728 & 0.82 & 0.852 & 0.687 & 0.853 \\
    (2) vs (3) & 0.825 & 0.833 & 0.753 & 0.713 & 0.81 & 0.849 & 0.726 & 0.842 \\
    (2) vs (4) & 0.931 & 0.93 & 0.736 & 0.812 & 0.943 & 0.927 & 0.704 & 0.944 \\
    (2) vs (5) & 0.967 & 0.962 & 0.761 & 0.851 & 0.973 & 0.96 & 0.735 & 0.973 \\
    (2) vs (6) & 0.818 & 0.825 & 0.743 & 0.706 & 0.805 & 0.84 & 0.679 & 0.838 \\
    (3) vs (4) & 0.83 & 0.842 & 0.691 & 0.689 & 0.817 & 0.862 & 0.69 & 0.847 \\
    (3) vs (5) & 0.815 & 0.831 & 0.691 & 0.682 & 0.806 & 0.852 & 0.696 & 0.837 \\
    (3) vs (6) & 0.993 & 0.995 & 0.853 & 0.952 & 0.979 & 0.996 & 0.816 & 0.995 \\
    (4) vs (5) & 0.964 & 0.971 & 0.952 & 0.954 & 0.962 & 0.974 & 0.954 & 0.972 \\
    (4) vs (6) & 0.83 & 0.842 & 0.782 & 0.735 & 0.822 & 0.861 & 0.787 & 0.848 \\
    (5) vs (6) & 0.814 & 0.83 & 0.759 & 0.716 & 0.811 & 0.85 & 0.761 & 0.837 \\
    \end{tabular}
\end{table*}


\end{document}